\pdfoutput=1
\RequirePackage{fix-cm}
\documentclass[natbib, letterpaper, 16pt]{svjour3_1}       
\smartqed  
\usepackage{graphicx}

\usepackage{mathptmx}      

\usepackage{amsmath, amssymb}
\usepackage{bm}
\usepackage{mathrsfs}
\usepackage{graphicx, fancyhdr}
\usepackage{placeins}  
\usepackage{floatrow}

\usepackage[intoc]{nomencl}

\usepackage{hyperref}

\let\origfontsize\fontsize
\def\fontsize#1#2{\origfontsize{12}{14.5}}

\hypersetup{
 colorlinks=true,
 linkcolor=blue,
 citecolor=blue,
 urlcolor=blue
 }

\usepackage{epstopdf}
\usepackage{color}
\usepackage[boxruled,vlined,linesnumbered]{algorithm2e}
\usepackage{ifpdf}
\usepackage{multirow}
\usepackage{caption}
\usepackage{subcaption}

\newdimen\figrasterwd
\figrasterwd\textwidth

\usepackage{setspace}
\usepackage{xcolor}

\usepackage[margin=1in]{geometry}

\newlength\myindent
\begin{document}
\title{3D CNN-PCA: A Deep-Learning-Based Parameterization for Complex Geomodels 
}

\titlerunning{Deep-learning-based geological parameterization}        

\author{Yimin Liu, Louis J.~Durlofsky}

\institute{Y. Liu \at
              Department of Energy Resources Engineering, Stanford University, \\
              Stanford, CA 94305-2220, USA \\
              \email{yiminliu@stanford.edu}           
           \and
           L.J. Durlofsky \at
              Department of Energy Resources Engineering, Stanford University, \\
              Stanford, CA 94305-2220, USA \\
              \email{lou@stanford.edu}           
}

\maketitle

\begin{abstract}

Geological parameterization enables the representation of geomodels in terms of a relatively small set of variables. Parameterization is therefore very useful in the context of data assimilation and uncertainty quantification. In this study, a deep-learning-based geological parameterization algorithm, CNN-PCA, is developed for complex 3D geomodels. CNN-PCA entails the use of convolutional neural networks as a post-processor for the low-dimensional principal component analysis representation of a geomodel. The 3D treatments presented here differ somewhat from those used in the 2D CNN-PCA procedure. Specifically, we introduce a new supervised-learning-based reconstruction loss, which is used in combination with style loss and hard data loss. The style loss uses features extracted from a 3D CNN pretrained for video classification. The 3D CNN-PCA algorithm is applied for the generation of conditional 3D realizations, defined on $60\times60\times40$ grids, for three geological scenarios (binary and bimodal channelized systems, and a three-facies channel-levee-mud system). CNN-PCA realizations are shown to exhibit geological features that are visually consistent with reference models generated using object-based methods. Statistics of flow responses ($\text{P}_{10}$, $\text{P}_{50}$, $\text{P}_{90}$ percentile results) for test sets of 3D CNN-PCA models are shown to be in consistent agreement with those from reference geomodels. Lastly, CNN-PCA is successfully applied for history matching with ESMDA for the bimodal channelized system.


\end{abstract}

\newcommand{\Ud}{\mathrm{d}} 
\newcommand{\Bx}{\mathbf{x}}
\newcommand{\By}{\mathbf{y}} 
\newcommand{\Bh}{\mathbf{h}} 
\newcommand{\Ba}{\mathbf{a}}
\newcommand{\Bb}{\mathbf{b}}
\newcommand{\Bxstar}{\mathbf{x}^{*}}
\newcommand{\Bz}{\mathbf{z}}
\newcommand{\Bxi}{\boldsymbol{\xi}}
\newcommand{\Bxih}{\hat{\boldsymbol{\xi}}}
\newcommand{\Bxit}{\Tilde{\boldsymbol{\xi}}}
\newcommand{\Beps}{\boldsymbol{\epsilon}}
\newcommand{\BXi}{\boldsymbol{\Xi}}
\newcommand{\Bximap}{\boldsymbol{\xi}_\text{map}}
\newcommand{\Bxirml}{\boldsymbol{\xi}_\text{rml}}
\newcommand{\Bxiuc}{\boldsymbol{\xi}^*}
\newcommand{\Bm}{\mathbf{m}_\text{gm}}
\newcommand{\Bw}{\mathbf{w}}
\newcommand{\BM}{\mathbf{M}}
\newcommand{\Bg}{\mathbf{g}}
\newcommand{\Bmref}{\Bm_{\text{ref}}}

\newcommand{\Bmbar}{\bar{\mathbf{m}}_\text{gm}}
\newcommand{\Bmuc}{\mathbf{m}_\text{uc}}
\newcommand{\Bmpca}{\mathbf{m}_\text{pca}}
\newcommand{\Bmpcah}{\hat{\mathbf{m}}_\text{pca}}
\newcommand{\Bmpcat}{\Tilde{\mathbf{m}}_\text{pca}}
\newcommand{\Bmopca}{\mathbf{m}_\text{opca}}
\newcommand{\Bmcnnpca}{\mathbf{m}_\text{cnnpca}}
\newcommand{\Bu}{\mathbf{u}}
\newcommand{\argmin}[1]{\underset{#1}{\text{argmin}}}
\newcommand{\erf}[1]{\text{erf}(#1)}
\newcommand{\Bd}{\mathbf{d}} 
\newcommand{\Bdobs}{\mathbf{d}_{\text{obs}}}
\newcommand{\Cd}{C_{\text{d}}}
\newcommand{\Cdinv}{C^{-1}_{\text{d}}}
\newcommand{\Cm}{C_{\text{m}}}
\newcommand{\Cminv}{C^{-1}_{\text{m}}}
\newcommand{\Nr}{N_{\text{r}}}
\newcommand{\Nx}{N_x}
\newcommand{\Ny}{N_y}
\newcommand{\Nz}{N_z}
\newcommand{\Nb}{N_{\text{b}}}
\newcommand{\Nc}{N_{\text{c}}}
\newcommand{\Nt}{N_{\text{t}}}
\newcommand{\Nh}{N_{\text{h}}}
\newcommand{\Nep}{N_\text{ep}}
\newcommand{\Nzz}[1]{N_{z,#1}}
\newcommand{\Nct}{N_{\text{c}}^{\text{t}}}
\newcommand{\td}{\text{d}}
\newcommand{\Bnu}{\boldsymbol{\nu}}
\newcommand{\COtwo}{\text{CO}_2}
\newcommand{\Smap}{S_\text{map}}
\newcommand{\Srml}{S_\text{rml}}
\newcommand{\Sd}{S_\text{d}}
\newcommand{\Sm}{S_\text{m}}
\newcommand{\lr}{l_\text{r}}

\newcommand{\Swir}{S_\text{wir}}
\newcommand{\Sor}{S_\text{or}}
\newcommand{\nw}{{n_\text{w}}}
\newcommand{\no}{{n_\text{o}}}
\newcommand{\Bmkr}{\mathbf{m}_\text{kr}}
\newcommand{\Bmkrbar}{\bar{\mathbf{m}}_\text{kr}}
\newcommand{\Bmkruc}{\mathbf{m}_\text{kr,uc}}
\newcommand{\Ckr}{C_\text{kr}}
\newcommand{\Ckrinv}{C_\text{kr}^{-1}}
\newcommand{\cmpd}{\text{m}^3/\text{day}}
\newcommand{\StdGauss}{N(\boldsymbol{0},I)}
\newcommand{\Nrml}{N_\text{rml}}
\newcommand{\Niter}{N_\text{iter}}
\newcommand{\Nd}{N_\text{d}}
\newcommand{\tsim}{t_\text{sim}}
\newcommand{\textapprox}{\raisebox{-0.6ex}{\textasciitilde}\hspace{0.15em}}
\newcommand{\BBR}{\mathbb{R}}

\newcommand{\mref}{M_\text{ref}}

\newcommand{\hp}{\hspace{8px}}
\newcommand{\hps}{\hspace{4px}}

\newcommand{\Lt}{L_\text{t}}
\newcommand{\Lc}{L_\text{c}}
\newcommand{\Ls}{L_\text{s}}
\newcommand{\Lh}{L_\text{h}}
\newcommand{\al}{\alpha_l}
\newcommand{\bl}{\beta_l}
\newcommand{\Nl}{N_l}
\newcommand{\Dl}{D_l}
\newcommand{\FO}{F_l[O]}
\newcommand{\FI}{F_l[I]}
\newcommand{\Fd}{F_l[\cdot]}
\newcommand{\GO}{G_l[O]}
\newcommand{\Gd}{G_l[\cdot]}
\newcommand{\GS}{G_l[S]}

\newcommand{\TI}{\text{TI}}

\newcommand{\beginsupplement}{%
        \setcounter{table}{0}
        \renewcommand{\thetable}{S\arabic{table}}%
        \setcounter{figure}{0}
        \renewcommand{\thefigure}{S\arabic{figure}}%
        \setcounter{section}{0}
        \renewcommand{\thesection}{S.\arabic{section}}%
     }

\section{Introduction}

In subsurface flow problems, data assimilation (history matching) typically entails the modification of geological parameters, such as permeability and porosity, such that flow simulation results are in essential agreement with observed data. Geological parameterization is often a useful component of this workflow, as it enables the mapping of high-dimensional (grid-block level) descriptions to a set of uncorrelated low-dimensional variables. A key challenge for parameterization procedures is the preservation of the large-scale geological structures that characterize the underlying geomodel. Recently, deep-learning-based parameterization algorithms have achieved promising results for the modeling of complex fluvial channel systems, though much of this work has been for 2D systems. Our goal in this study is to extend one such parameterization, CNN-PCA (convolutional neural network -- principal component analysis), to handle complex 3D geomodels.




Deep-learning-based treatments are now being applied for many aspects of subsurface characterization and flow modeling. This includes, e.g., surrogate modeling for flow \citep{Mo2019, Wen2019, Tang2020, Jin2020}, data-space inversion for history matching and uncertainty quantification \citep{Jiang2020}, and dimension reduction of seismic data in time-lapse seismic history matching \citep{Liu2020b,Liu2020a}. Initial work on deep-learning-based geological parameterization includes formulations involving fully connected autoencoders (AE) \citep{Canchumuni2017} and deep belief networks (DBN) \citep{Canchumuni2018}. CNN-based algorithms are, in general, more efficient and scalable. Examples involving CNNs include convolutional variational autoencoders (VAE) \citep{Laloy2017, Canchumuni2019a} and deep convolutional generative adversarial networks (GAN) \citep{Chan2017, Chan2018, Dupont2018, Mosser2018, Laloy2018, Laloy2019, Chan2020, Azevedo2020}. The combination of VAE and GAN has also been proposed by \cite{Mo2020} and \cite{Canchumun2020}. 

The above-mentioned algorithms utilize deep-learning to establish an end-to-end mapping between geological models and low-dimensional latent variables. The CNN-PCA \citep{Liu2019,Liu2020} and PCA-Cycle-GAN \citep{Canchumun2020} algorithms take a different approach, in that deep-learning is used as a post-processor for PCA, which is a traditional parameterization method. In 2D CNN-PCA, a style loss based on geomodel features extracted from the VGG net \citep{Simonyan2015a} is employed to improve geological realism. This idea has also been employed to improve the quality of parameterized models from a standard VAE \citep{Canchumun2020}. A wide range of deep-learning-based geological parameterization algorithms have been developed and tested on 2D systems. They have been shown to provide more realistic models than traditional methods such as standalone PCA \citep{Sarma2006} and discrete cosine transform \citep{Jafarpour2010}. In a benchmark comparison performed by \cite{Canchumun2020}, seven deep-learning-based parameterization techniques, including VAE-based, GAN-based and PCA-based algorithms, were found to perform similarly for a 2D channel system.  



The development and detailed testing of deep-learning-based parameterizations for 3D geomodels is much more limited. Most assessments involving 3D systems are limited to unconditional realizations of a single geological scenario \citep{Laloy2018, Canchumuni2018, Mo2020}. To our knowledge, conditional realizations of 3D geomodels have thus far only been considered by \cite{Laloy2017} for the Maules Creek Valley alluvial aquifer dataset \citep{ti_lib}. Thus there appears to be a need for work on deep-learning-based parameterizations for complex 3D geomodels.

In this study, we extend the 2D CNN-PCA framework to 3D by incorporating several new treatments. These include the use of a supervised-learning-based training loss and the replacement of the VGG net (appropriate for 2D models) with the C3D net \citep{Tran2015}, pretrained for video classification, as a 3D feature extractor. This formulation represents a significant extension of the preliminary treatment presented by \cite{TangLiu2020}, where style loss was not considered (in that work geomodel parameterization was combined with surrogate models for flow in a binary system). Here we apply the extended 3D CNN-PCA procedure for the parameterization of conditional realizations within three different geological settings. These include a binary fluvial channel system, a bimodal channel system, and a three-facies channel-levee-mud system. In all cases the underlying realizations, which provide the training sets for the 3D CNN-PCA parameterizations, are generated using object-based modeling within the Petrel geomodeling framework \citep{manual2007petrel}. Geomodels and results for flow statistics are presented for all three of the geological scenarios, and history matching results are presented for the bimodal channel case.



This paper proceeds as follows. In Section~\ref{sec-methodology}, we begin by briefly reviewing the existing (2D) CNN-PCA algorithm. The 3D CNN-PCA methodology is then discussed in detail. In Section~\ref{sec-model-gen}, we apply 3D CNN-PCA to parameterize facies models for a binary channel system and a three facies channel-levee-mud system. A two-step approach for treating bimodal systems is then presented. For all of these geological scenarios, 3D CNN-PCA realizations, along with flow results for an ensemble of test cases, are presented. In Section~\ref{sec-hm}, we present history matching for the bimodal channel system using the two-step parameterization and ESMDA. A summary and suggestions for future work are provided in Section~\ref{sec-concl}. Details on the network architectures, along with additional geomodeling and flow results, are included in Supplementary Information (SI).

\section{Convolutional Neural Network-based Principal Component Analysis (CNN-PCA)}
\label{sec-methodology}

In this section, we first give a brief overview of PCA and the 2D CNN-PCA method. The 3D procedure is then introduced and described in detail.

\subsection{PCA Representation}
\label{sec-pca}

We let the vector $\mathbf{m} \in \mathbb{R}^{\Nc}$, where $\Nc$ is the number of cells or grid blocks, denote the set of geological variables (e.g., facies type in every cell) that characterize the geomodel. Parameterization techniques map $\mathbf{m}$ to a new lower-dimensional variable $\Bxi \in \mathbb{R}^{l}$, 
where $l < \Nc$ is the reduced dimension. As discussed in detail in \cite{Liu2019}, PCA applies linear mapping of $\mathbf{m}$ onto a set of  principal components. To construct a PCA representation, an ensemble of $\Nr$ models is generated using a geomodeling tool such as Petrel \citep{manual2007petrel}. These models are assembled into a centered data matrix $Y \in \mathbb{R}^{\Nc \times \Nr}$,
\begin{equation}
    \label{eq_center_data_matrix}
    Y = \frac{1}{\sqrt{\Nr - 1}}[\Bm^1 - \Bmbar \quad \Bm^2 - \Bmbar \quad \cdots \quad \Bm^{\Nr} - \Bmbar],
\end{equation}
where $\Bm^i\in \mathbb{R}^{\Nc}$ represents realization $i$, $\Bmbar\in \mathbb{R}^{\Nc}$ is the mean of the $\Nr$ realizations, and the subscript `gm' indicates that these realizations are generated using geomodeling software. A singular value decomposition of $Y$ gives $Y = U\Sigma V^T$, where $U \in \mathbb{R}^{\Nc \times \Nr}$ and $V \in \mathbb{R}^{\Nr \times \Nr}$ are the left and right singular matrices and $\Sigma \in \mathbb{R}^{\Nr \times \Nr}$ is a diagonal matrix containing singular values. 

A new PCA model $\Bmpca \in \mathbb{R}^{\Nc}$ can be generated as follows,
\begin{equation}
\label{eq_pca}
    \Bmpca = \Bmbar + U_l\Sigma_l \Bxi_l,
\end{equation}
where $U_l \in \mathbb{R}^{\Nc \times l}$ and $\Sigma_l \in \mathbb{R}^{l \times l}$ contain the leading left singular vectors and singular values, respectively. Ideally, it is the case that $l << \Nc$. 
By sampling each component of $\Bxi_l$ independently from the standard normal distribution and applying Eq.~\ref{eq_pca}, we can generate new PCA models.

Besides generating new models, we can also apply PCA to approximately reconstruct realizations of the original models. We will see later that this is required for the supervised-learning-based loss function used to train 3D CNN-PCA. Specifically, we can project each of the $\Nr$ realizations of $\Bm$ onto the principal components via
\begin{equation}
\label{eq_pca_proj}
    \Bxih_l^i = \Sigma^{-1}_lU^T_l(\Bm^i - \Bmbar), \hp i=1,...,\Nr .
\end{equation}
Here $\Bxih_l^i$ denotes low-dimensional variables obtained through projection. The `hat' is added to differentiate $\Bxih_l^i$ from low-dimensional variables obtained through sampling ($\Bxi_l$). We can then approximately reconstruct $\Bm^i$ as
\begin{equation}
\label{eq_pca_recon}
    \Bmpcah^i = \Bmbar + U_l \Sigma_l \Bxih_l^i = \Bmbar + U_l U_l^T (\Bm^i - \Bmbar), \hp i=1,...,\Nr,
\end{equation}
where $\Bmpcah^i$ is referred to as a reconstructed PCA model. The larger the reduced dimension $l$, the closer $\Bmpcah^i$ will be to $\Bm^i$. If all $\Nr - 1$ nonzero singular values are retained, $\Bmpcah^i$ will exactly match $\Bm^i$. 

For systems where $\Bm$ follows a multi-Gaussian distribution, the spatial correlation of $\Bm$ can be fully characterized by two-point correlations, i.e., covariance. For such systems, $\Bmpca$ (constructed using Eq.~\ref{eq_pca}) will essentially preserve the spatial correlations in $\Bm$, assuming $l$ is sufficiently large. For systems with complex geology, where $\Bm$ follows a non-Gaussian distribution, the spatial correlation of $\Bm$ is characterized by multiple-point statistics. In such cases, the spatial structure of $\Bmpca$ can deviate significantly from that of $\Bm$, meaning the direct use of Eq.~\ref{eq_pca} is not appropriate.

\subsection{2D CNN-PCA Procedure}
\label{Subsect_cnn_pca}

In CNN-PCA, we post-process PCA models using a deep convolutional neural network to achieve better correspondence with the underlying geomodels. This process can be represented as 
\begin{equation}
    \Bmcnnpca = f_W(\Bmpca),
\end{equation}
where $f_W$ denotes the model transform net, the subscript $W$ indicates the trainable parameters within the network, and $\Bmcnnpca \in \mathbb{R}^{\Nc}$ is the resulting geomodel.

For 2D models, the training loss ($L^i$) for $f_W$, for each training sample $i$, includes a content loss ($\Lc^i$) and a style loss ($\Ls^i$):
\begin{equation}
\label{eq:2Dloss}
    L^i = \Lc^i(f_W(\Bmpca^i), \Bmpca^i) + \Ls^i(f_W(\Bmpca^i), \mref), \hp i=1,..,\Nt,
\end{equation}
where $\Bmpca^i, i=1,..,\Nt$ is a training set (of size $\Nt$) of random new PCA models, and $\mref$ is a reference model (e.g., training image or an original realization $\Bm$). Here $\Lc^i$ quantifies the `closeness' of $f_W(\Bmpca^i)$ to $\Bmpca^i$, and acts to ensure that the post-processed model resembles (to some extent) the input PCA model. The style loss $\Ls^i$ quantifies the resemblance of $f_W(\Bmpca^i)$ to $\mref$ in terms of spatial correlation structure.

As noted in Section~\ref{sec-pca}, the spatial correlation of non-Gaussian models can be characterized by high-order multiple-point statistics. It is not practical, however, to compute such quantities directly. Therefore, $\Ls^i$ in Eq.~\ref{eq:2Dloss} is not based on high-order spatial statistics but rather on low-order statistics of features extracted from another pretrained CNN, referred to as the loss net \citep{Gatys2015, Johnson2016}. More specifically, we feed the 2D models $\Bm$, $\Bmpca$ and $f_W(\Bmpca)$ through the loss net and extract intermediate feature matrices $F_k(\mathbf{m})$ from different layers $k \in \kappa$ of the loss net. The uncentered covariance matrices, called Gram matrices, are given by $G_k(\mathbf{m}) = F_k(\mathbf{m})F_k(\mathbf{m})^T/(N_{\text{c},k}\Nzz{k})$, where $N_{\text{c},k}$ and $\Nzz{k}$ are the dimensions of $F_k(\mathbf{m})$. These matrices have been shown to provide an effective set of metrics for quantifying the multipoint correlation structure of 2D models. The style loss is thus based on the differences between $f_W(\Bmpca^i)$ and reference model $\mref$ in terms of their corresponding Gram matrices. This is expressed as 
\begin{equation}
\label{eq-ls}
\Ls^i(f_W(\Bmpca^i), \mref) = \sum_{k \in \kappa}\dfrac{1}{\Nzz{k}^2}||G_k(f_W(\Bmpca^i)) - G_k(\mref)||, \hp i=1,..,\Nt.
\end{equation}
%
The content loss is based on the difference between the feature matrices for $f_W(\Bmpca^i)$ and $\Bmpca^i$ from a particular layer in the network.

For the 2D models considered in \cite{Liu2019} and \cite{Liu2020}, VGG net was used as the loss net \citep{Simonyan2015a}. We have a concept of transfer learning here as the VGG net, pretrained on image classification, was shown to be effective at extracting features from 2D geomodels. However, since VGG net only accepts image-like 2D input, it cannot be directly used for extracting features from 3D geological models. Our extension of CNN-PCA to 3D involves two main components: the replacement of VGG net with 3D CNN models, and the use of a new loss term based on supervised learning. We now describe the 3D CNN-PCA formulation.



\subsection{3D CNN-PCA Formulation}

We experimented with several pretrained 3D CNNs for extracting features from 3D geomodels, which are required to compute style loss in 3D CNN-PCA. These include VoxNet \citep{Maturana2015} and LightNet \citep{Zhi2017} for 3D object recognition, and C3D net \citep{Tran2015} for classification of video clips in the sports-1M dataset \citep{Karpathy2014}. These CNNs accept input as dense 3D tensors with either three spatial dimensions or two spatial dimensions and a temporal dimension. Thus all are compatible for use with 3D geomodels. After numerical experimentation with these various networks, we found the C3D net to perform the best, based on visual inspection of the geomodels generated by the trained model transform nets. Therefore, we use the C3D net in our 3D CNN-PCA formulation.

We observed, however, that the Gram matrices extracted from the C3D net were not as effective as they were with VGG net in 2D. This is likely due to the higher dimensionality and larger degree of variability of the 3D geomodels relative to those in 2D. We therefore considered additional treatments, and found that the use of a new supervised-learning-based loss term provides enhanced 3D CNN-PCA geomodels. We now describe this procedure.

As discussed in Section~\ref{sec-pca}, we can approximately reconstruct realizations of the original model $\Bm$ with PCA using Eq.~\ref{eq_pca_recon}. There is however reconstruction error between the two sets of models. The supervised learning component entails training the model transform net to minimize an appropriately defined reconstruction error. 

Recall that when the trained model transform net is used at test time (e.g., to generate new random models or to calibrate geomodels during history matching), this entails post-processing new PCA models $\Bmpca(\Bxi_l)$. Importantly, these new models involve $\Bxi_l$ rather than $\Bxih_l^i$. In other words, at test time we do not have corresponding pairs of $(\Bm^i, \Bmpcah^i)$. Thus, during training, it is beneficial to partially `disrupt' the direct correspondence that exists between each pair of $(\Bm^i, \Bmpcah^i)$. An effective way of accomplishing this is to perturb the reconstructed PCA models used in training. We proceed by adding random noise to the $\Bxih_l^i$ in Eq.~\ref{eq_pca_proj}; i.e.,
%
\begin{equation}
\Bxit_l^i = \Bxih_l^i +  \Beps^i = \Sigma^{-1}_lU^T_l(\Bm^i - \Bmbar) + \Beps^i, \hp i=1,...,\Nr,
\label{eq_pca_proj_perturb}
\end{equation}
where $\Bxit_l^i$ denotes the perturbed low-dimensional variable and $\Beps^i$ is a perturbation term. Then we can approximately reconstruct $\Bm^i$ with
\begin{equation}
\label{eq_pca_recon_perturb}
    \Bmpcat^i = \Bmbar + U_l\Sigma_l \Bxit_l^i, \hp i=1,...,\Nr.
\end{equation}

We now describe how we determine $l$ and specify $\Beps$. To find $l$, we apply the `energy' criterion described in \cite{Sarma2006} and \cite{Vo2014}. This entails first determining the total energy $E_\text{t} = \sum_{i=1}^{\Nr-1}(\sigma^i)^2$, where $\sigma^i$ are the singular values. The fraction of energy captured by the $l$ leading singular values is given by $\sum_{i=1}^{l}(\sigma^i)^2 / E_\text{t}$. Throughout this study, we determine $l$ such that the $l$ leading singular values explain \textapprox80\% of the total energy. For the components of the perturbation term $\Beps$, we set $\epsilon_j = 0$ for $j=1,...,p$, where $p$ is determined such that the first $p$ leading singular values explain \textapprox40\% of the total energy, and $\epsilon_j \mathtt{\sim} N(0,1)$ for $j=p+1,...,l$. With this treatment, we perturb only the small-scale features in $\Bmpcat$. This approach was found to be effective as it disrupts the precise correspondence between $(\Bm^i, \Bmpcah^i)$ pairs, while maintaining the locations of major geological features. 

We use the same $\Nr$ realizations of $\Bm$ as were used for constructing the PCA representation for the generation of $\Bmpcat$. We reiterate that Eqs.~\ref{eq_pca_proj_perturb} and \ref{eq_pca_recon_perturb} are used here (Eqs.~\ref{eq_pca_proj} and \ref{eq_pca_recon} are not applied). The supervised-learning loss function for each pair of $(\Bm^i, \Bmpcat^i)$, which we refer to as the reconstruction loss, is given by
\begin{equation}
\label{eq:3D_rec_loss}
    L_\text{rec}^i(\Bm^i, f_W(\Bmpcat^i)) = ||\Bm^i- f_W(\Bmpcat^i)||_1, \hp i=1,...,\Nr.
\end{equation}
Note that in 3D CNN-PCA we take $\Nt=\Nr$ in all cases.

The style loss is evaluated using a separate set of $\Nr$ new PCA models $\Bmpca^i(\Bxi_l^i)$, $i=1,...,\Nr$, with $\Bxi_l^i$ sampled from \textcolor{blue}{$N(\mathbf{0},I_l)$}. As for the reference model $\mref$, in 2D CNN-PCA we used either a reference training image or one realization of $\Bm$. Here we generate realizations of the original geomodel using object-based techniques in Petrel, so there is no reference training image. We therefore use realizations of $\Bm$ to represent the reference. Instead of using one particular realization, all $\Nr$ realizations of $\Bm$ (in turn) are considered as reference models. Specifically, we use $\Bm^i$ as the reference model for new PCA model $\Bmpca^i$. It is important to emphasize that $\Bm^i$ and $\Bmpca^i$ are completely unrelated in terms of the location of geological features -- we are essentially assigning a random reference model ($\Bm^i$) for each $\Bmpca^i$. However, because the style loss is based on summary spatial statistics, the exact location of geological features does not affect the evaluation of the loss. 

The style loss between $\Bm^i$ and the (non-corresponding) new PCA model $\Bmpca^i$ is given by
\begin{equation}
\label{eq:3D_style_loss}
    L_\text{s}^i(\Bm^i, f_W(\Bmpca^i)) = \sum_{k \in \kappa}\dfrac{1}{\Nzz{k}^2}||G_k(\Bm^i) - G_k(f_W(\Bmpca^i))||_1, \hp i=1,...,\Nr,
\end{equation}
where $G_k$ are Gram matrices based on features extracted from different layers in the C3D net. The C3D net consists of four blocks of convolutional and pooling layers. Here we use the last convolutional layer of each block, which corresponds to $k= 1, 2, 4, 6$. Details on the network architecture are provided in SI.

A hard data loss term is also include to assure hard data (e.g., facies type at well locations) are honored. Hard data loss $\Lh^i$ is given by
\begin{equation}
    \Lh^i = \dfrac{1}{\Nh}\left[\Bh^T(\Bm^i-f_W(\Bmpca^i))^2 + \Bh^T(\Bm^i-f_W(\Bmpcat^i))^2\right], \hp i=1,...,\Nr,
\end{equation}
where $\Bh$ is a selection vector, with $h_j=1$ indicating the presence of hard data at cell $j$ and $h_j=0$ the absence of hard data, and $\Nh$ is the total number of hard data. The final training loss is a weighted combination of the reconstruction loss, style loss and hard data loss. For each pair of corresponding $(\Bm^i, \Bmpcat^i)$, and the unrelated new PCA model $\Bmpca^i$, the total training loss is thus
\begin{equation}
\label{eq_cnnpca_loss}
L^i = \gamma_r L_\text{rec}^i(\Bm^i, f_W(\Bmpcat^i)) + \gamma_s L_\text{s}^i(\Bm^i, f_W(\Bmpca^i)) + \gamma_h \Lh^i, \hp i=1,...,\Nr.
\end{equation}
The three weighting factors $\gamma_r$, $\gamma_s$ and $\gamma_h$ are determined heuristically by training the network with a range of values and selecting the combination that leads to the lowest mismatch in quantities of interest (here we consider flow statistics) relative to the original (Petrel) geomodels. We also require that at least 99.9\% of the hard data are honored over the entire set of $\Bmcnnpca$ models. The training set is divided into multiple mini-batches, and the total loss for each mini-batch of samples is 
\begin{equation}
    \label{eq_cnnpca_loss_total}
    \Lt = \sum_{i=1}^{\Nb} L^i,
\end{equation}
where $\Nb$ is the batch size.

\begin{figure}[!htb]
    \centering
    \includegraphics[width=1\textwidth]{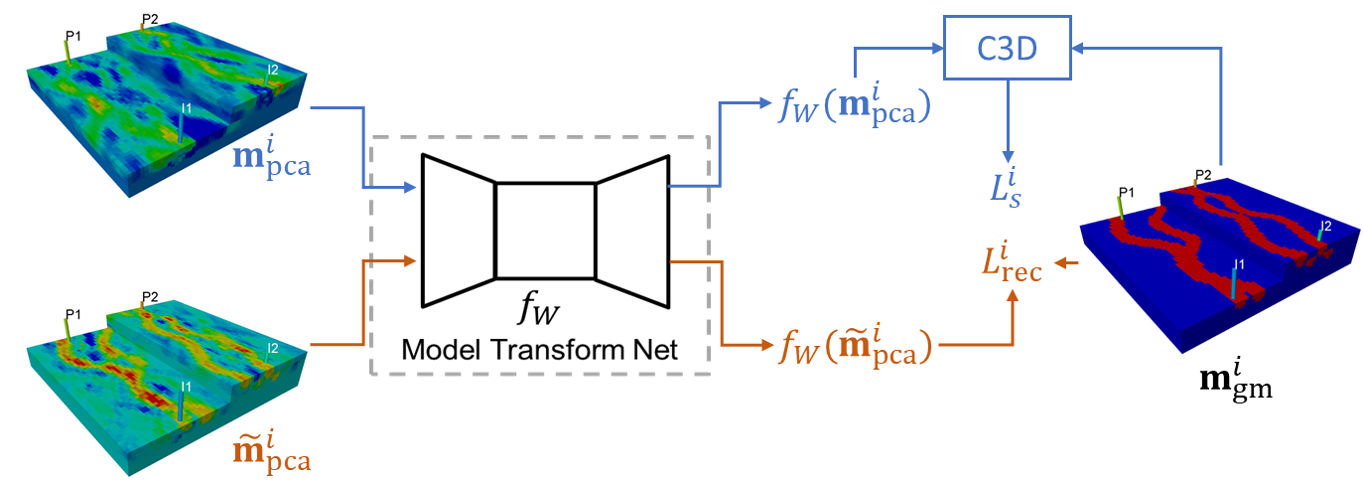}
    \caption{Training procedure for 3D CNN-PCA.}
    \label{fig-cnncpa-train}
\end{figure}

The model transform net for 3D CNN-PCA is obtained by replacing the 2D convolutional layers, upsampling layers, downsampling layers and padding layers in 2D CNN-PCA \citep{Liu2019} with their 3D counterparts. This can be readily accomplished within the Pytorch deep-learning framework \citep{Paszke2017}. The training procedure for 3D CNN-PCA is illustrated in Fig.~\ref{fig-cnncpa-train}. Each training sample consists of a pair of corresponding models $(\Bm^i, \Bmpcat^i)$ and an unrelated new PCA model $\Bmpca^i$. The new PCA model $\Bmpca^i$ and the reconstructed PCA model $\Bmpcat^i$ are fed through the model transform net $f_W$. The reconstruction loss is evaluated using $f_W(\Bmpcat^i)$ and the original model $\Bm^i$ and Eq.~\ref{eq:3D_rec_loss}. To evaluate the style loss, $f_W(\Bmpca^i)$ and $\Bm^i$ are fed through the C3D net, and the relevant feature matrices are extracted. Then, Gram matrices are computed to form the style loss (Eq.~\ref{eq:3D_style_loss}). The final loss entails a weighted combination of reconstruction loss, style loss and hard data loss. The trainable parameters in $f_W$ are updated based on the gradient of the loss computed with back-propagation. This process iterates over all mini-batches and continues for a specified number of epochs. The detailed architectures for the model transform and C3D nets are provided in SI.

\section{Geomodels and Flow Results Using 3D CNN-PCA}
\label{sec-model-gen}

We now apply the 3D CNN-PCA procedure to generate geomodels corresponding to three different geological scenarios (binary channelized, three-facies, and bimodal channelized systems). Visualizations of the CNN-PCA models, along with results for key flow quantities, are presented. All flow simulations in this work are performed using Stanford's Automatic Differentiation General Purpose Research Simulator, ADGPRS \citep{zhou2012parallel}.

\subsection{Case 1 -- Binary Channelized System}

\begin{figure}[!htb]
    \centering
    \begin{subfigure}[b]{0.24\textwidth}
        \includegraphics[width=1\textwidth]{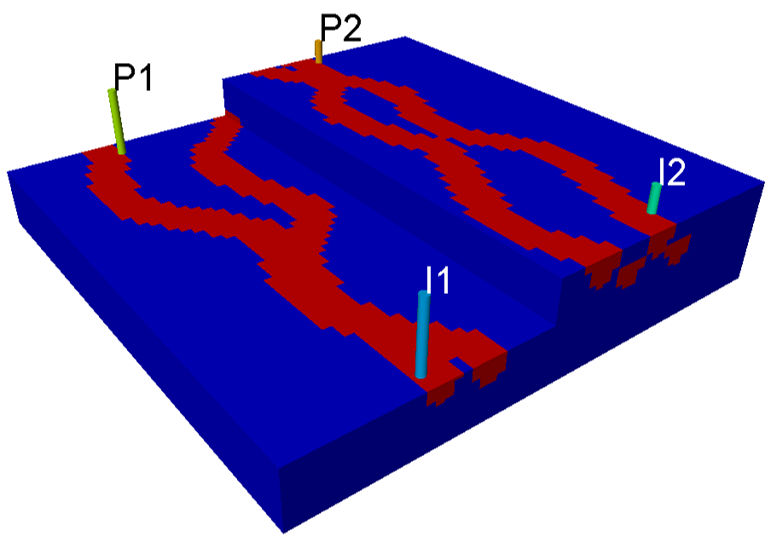}
    \end{subfigure}%
    ~ 
    \begin{subfigure}[b]{0.24\textwidth}
        \includegraphics[width=1\textwidth]{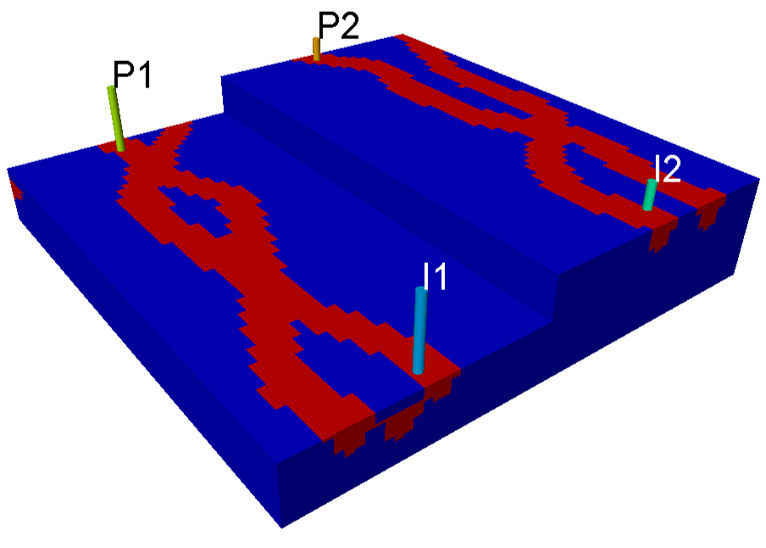}
    \end{subfigure}%
    ~
    \begin{subfigure}[b]{0.24\textwidth}
        \includegraphics[width=1\textwidth]{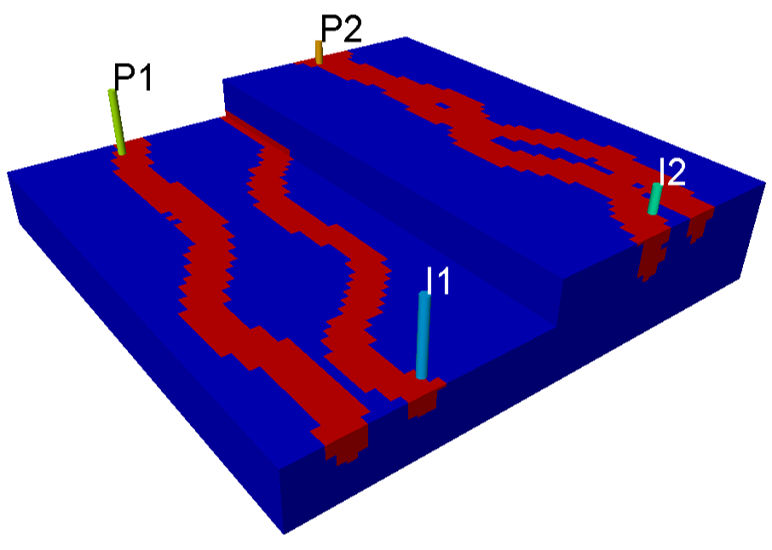}
    \end{subfigure}%
    ~
    \begin{subfigure}[b]{0.24\textwidth}
        \includegraphics[width=1\textwidth]{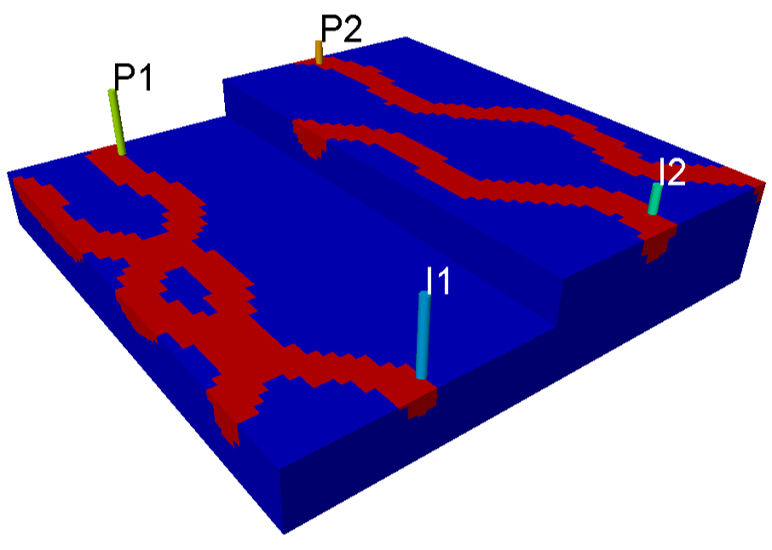}
    \end{subfigure}%
    \caption{Four realizations of the binary channel system generated using Petrel (Case~1).}
    \label{fig-chan-petrel}
\end{figure}

The first case involves a channelized system characterized by rock facies type, with 1 denoting high-permeability sandstone and 0 indicating low-permeability mud. The geomodels are defined on a $60\times60\times40$ grid (144,000 total cells). The average sand fraction is 6.53\%. Figure~\ref{fig-chan-petrel} displays four random facies realizations generated using object-based modeling within Petrel (sandstone is shown in red, and mud in blue).


\begin{figure}[!htb]
    \centering
    \begin{subfigure}[b]{1\textwidth}
        \includegraphics[width=0.24\textwidth]{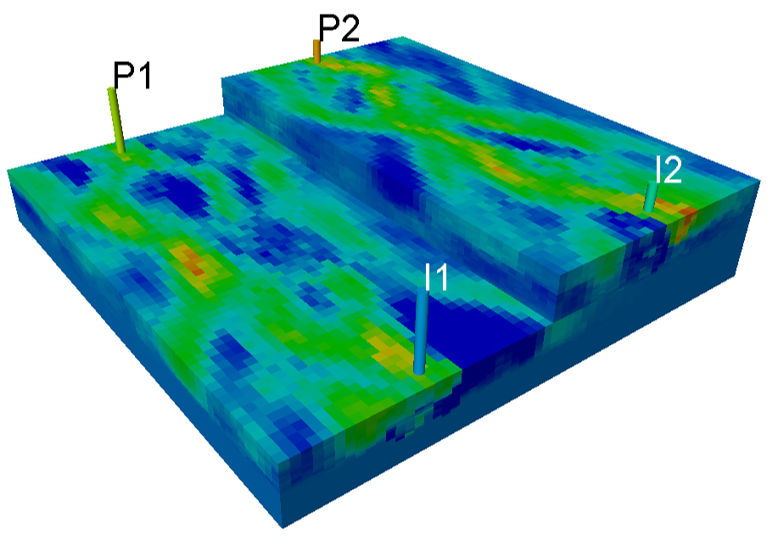}
        \includegraphics[width=0.24\textwidth]{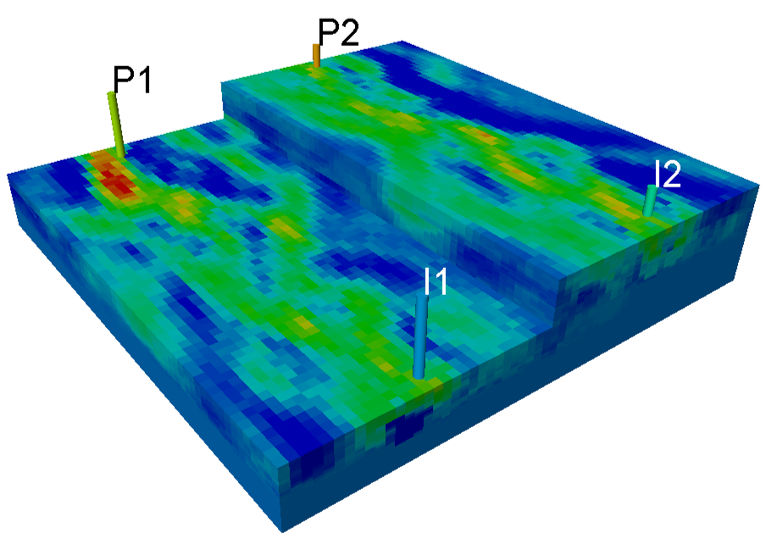}
        \includegraphics[width=0.24\textwidth]{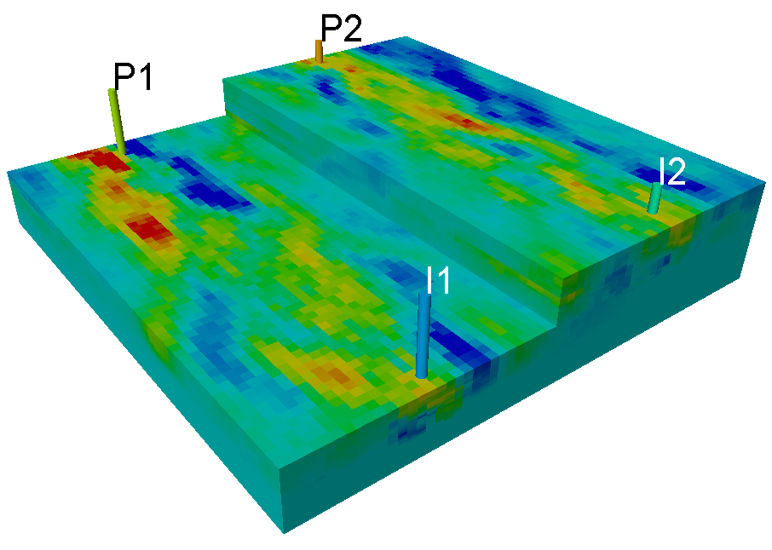}
        \includegraphics[width=0.24\textwidth]{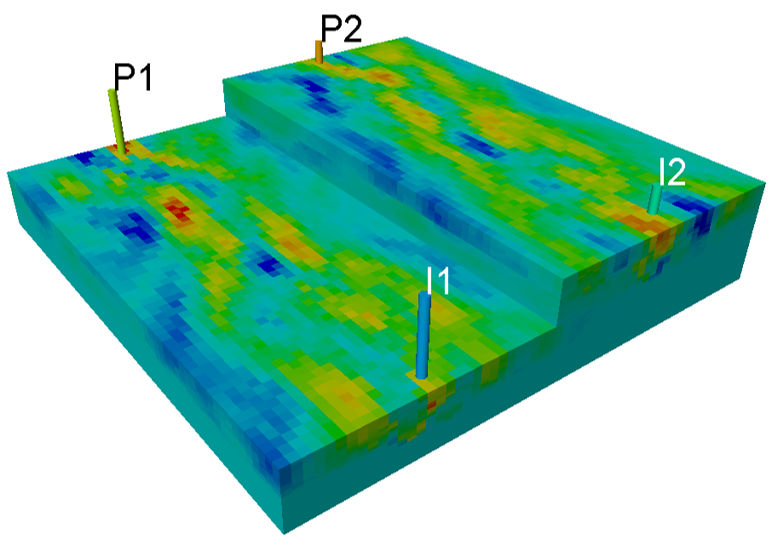}
        \caption{PCA models}
    \end{subfigure}%
    
    \begin{subfigure}[b]{1\textwidth}
        \includegraphics[width=0.24\textwidth]{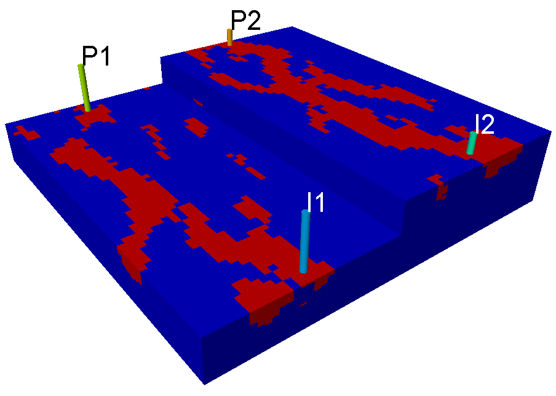}
        \includegraphics[width=0.24\textwidth]{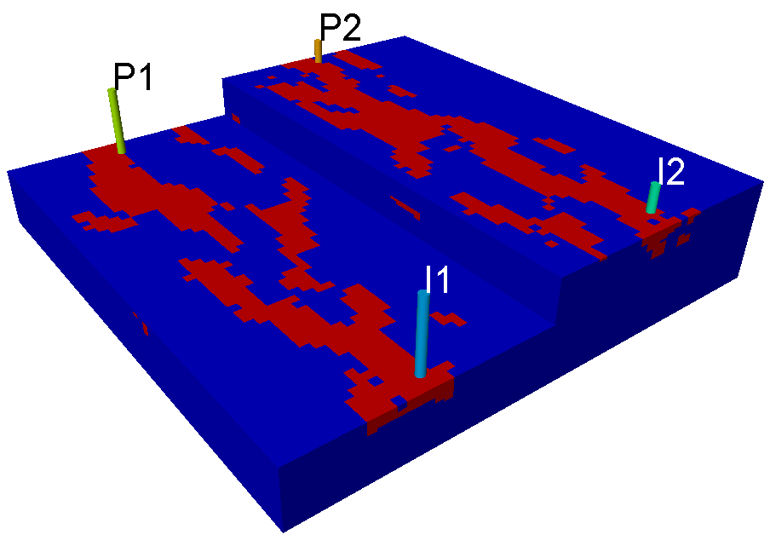}
        \includegraphics[width=0.24\textwidth]{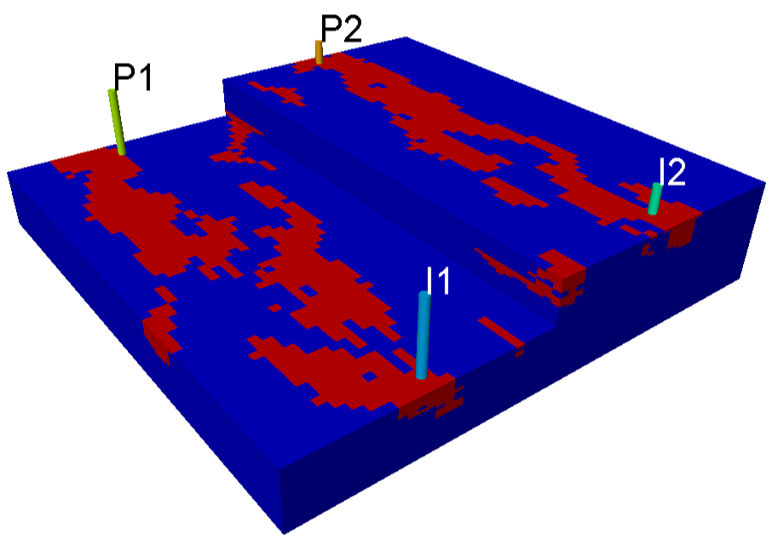}
        \includegraphics[width=0.24\textwidth]{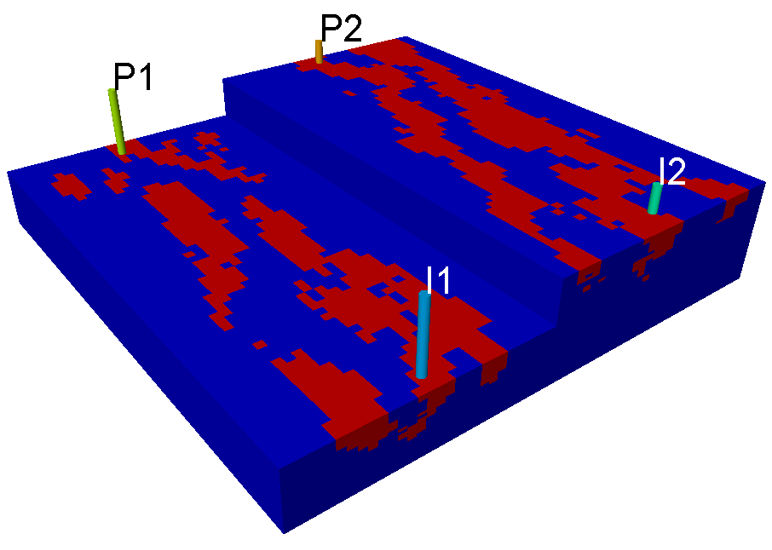}
        \caption{T-PCA models}
    \end{subfigure}%
    
    \begin{subfigure}[b]{1\textwidth}
        \includegraphics[width=0.24\textwidth]{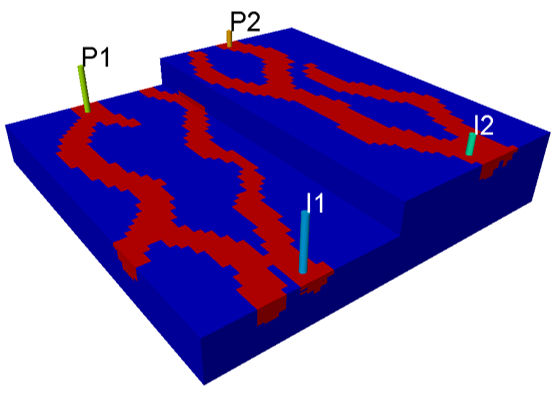}
        \includegraphics[width=0.24\textwidth]{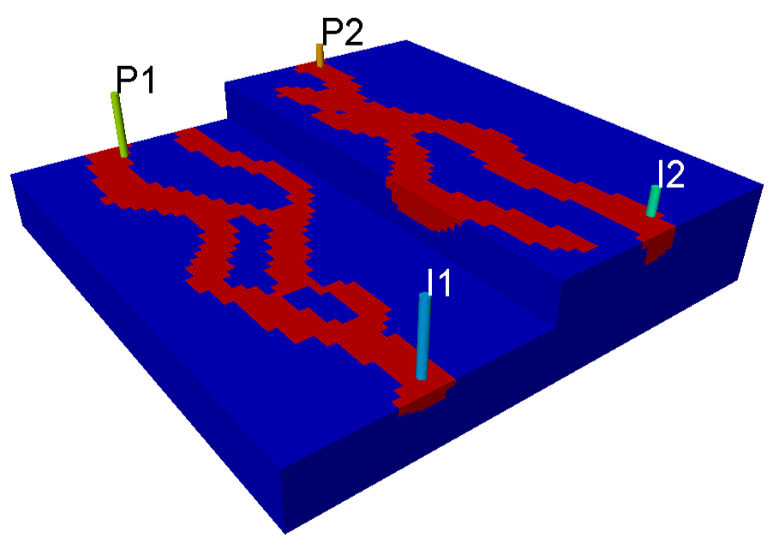}
        \includegraphics[width=0.24\textwidth]{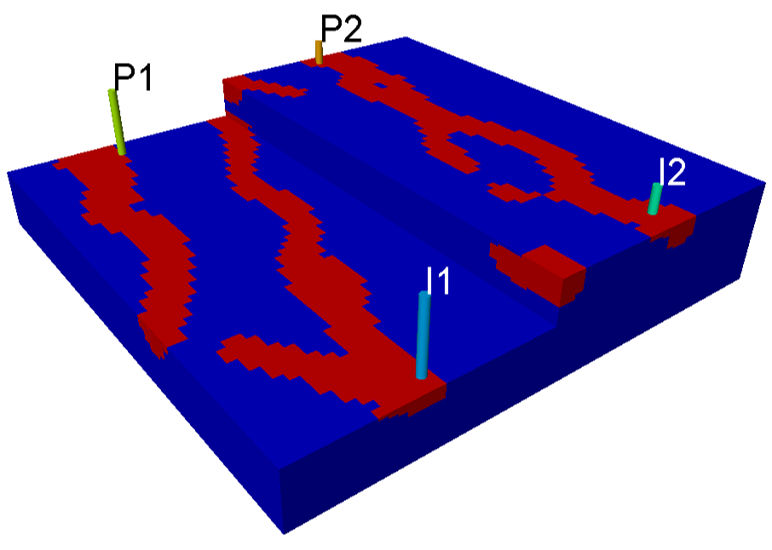}
        \includegraphics[width=0.24\textwidth]{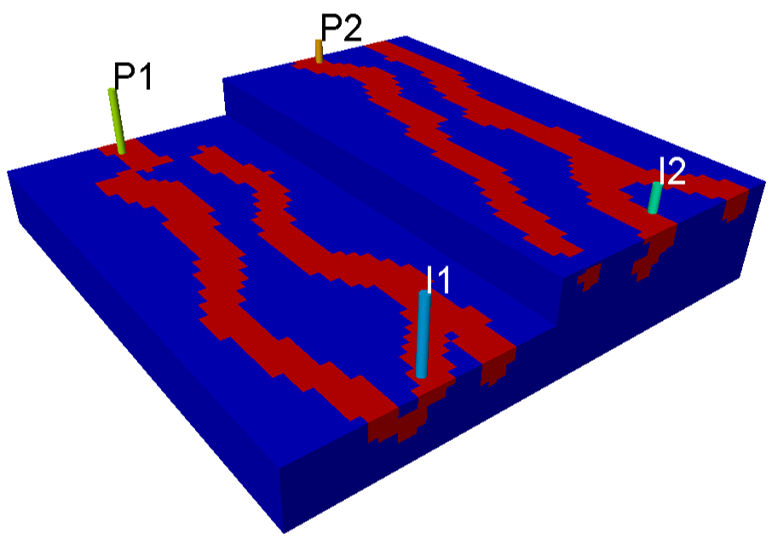}
        \caption{CNN-PCA models}
    \end{subfigure}%

    \caption{Four test-set realizations of the binary channel system from (a) PCA models, (b) corresponding truncated-PCA (T-PCA) models, and (c) corresponding CNN-PCA models (Case~1).}
    \label{fig_case1_models}
\end{figure}

In this case, there are two production wells and two injection wells. The well locations are given in Table~\ref{tab_well_case1}. All wells are assumed to be drilled through all 40~layers of the model, and hard data are specified (meaning $h_j=1$) in all blocks penetrated by a well. Wells are perforated (open to flow), however, only in blocks characterized by sand. These layers are indicated in Table~\ref{tab_well_case1}.


A total of $\Nr=3000$ conditional realizations $\Bm^i$ are generated to construct the PCA model (through application of Eq.~\ref{eq_center_data_matrix}). A total of $l=400$ singular values are retained, which explains $\sim$80\% of the total energy. Then, $\Nr=3000$ reconstructed PCA models $\Bmpcat^i$ are generated using Eqs.~\ref{eq_pca_proj_perturb} and \ref{eq_pca_recon_perturb}. The first $p=40$ principal components explain $\sim$40\% of the energy, so perturbation is only applied to $\xi_j$ for $j=41, ..., 400$. A separate set of $\Nr=3000$ random PCA models $\Bmpca^i$ is generated by sampling $\Bxi_l$ from $N(\mathbf{0},I_l)$. 

\begin{table}[!htb]
\centering
\begin{tabular}{ c | c | c | c |c  }
  & P1 & P2 & I1 & I2 \\
  \hline
  Areal location ($x,~y$)&(15, 57)&(45, 58)&(15, 2)&(45, 3)\\
  Perforated layers&18 - 24&1 - 8&15 - 22&1 - 8\\
\end{tabular}

\caption{Well locations ($x$ and $y$ refer to areal grid-block indices) and perforations (Case~1)}
\label{tab_well_case1}
\end{table}

The $\Nr=3000$ realizations of $\Bm^i$, $\Bmpcat^i$ and $\Bmpca^i$ form the training set for the training of the model transform net $f_W$. The weighting factors for the training loss in Eq.~\ref{eq_cnnpca_loss} are $\gamma_\text{rec}=500$, $\gamma_s=100$, and $\gamma_h = 10$. These values were found to provide accurate flow statistics and near-perfect hard-data honoring. The Adam optimizer \citep{Kingma2014} is used for updating parameters in $f_W$, with a default learning rate of $l_r=0.001$ and a batch size of $\Nb=8$. The model transform net is trained for 10~epochs, which requires around 0.5~hour on one Tesla V100 GPU.

\begin{figure}[!htb]
    \centering
    \begin{subfigure}[b]{0.32\textwidth}
        \includegraphics[width=1\textwidth]{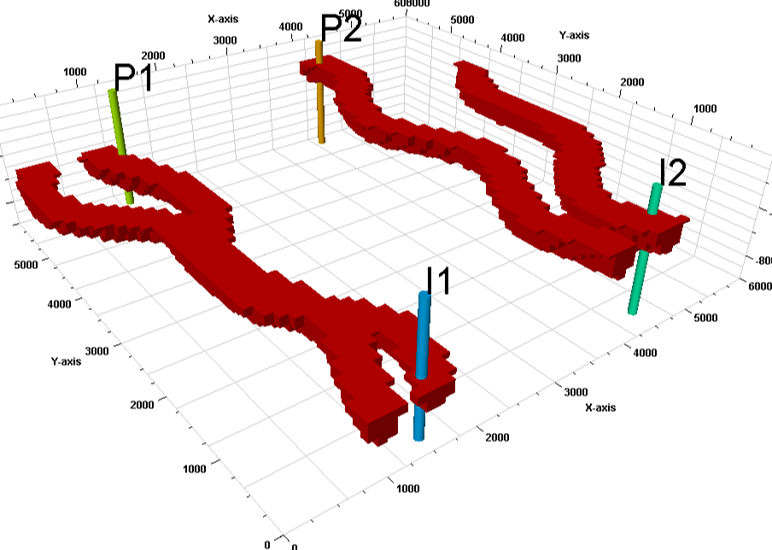}
        \caption{One Petrel model}
    \end{subfigure}%
    ~ 
    \begin{subfigure}[b]{0.32\textwidth}
        \includegraphics[width=1\textwidth]{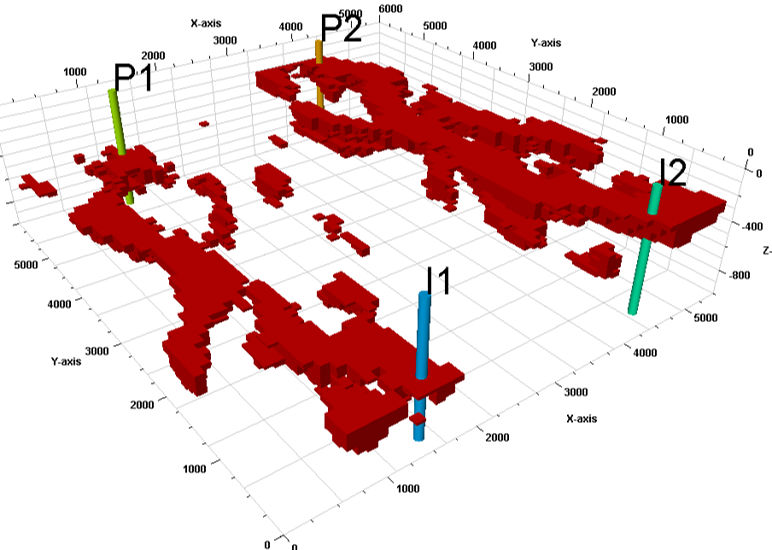}
        \caption{One T-PCA model}
    \end{subfigure}%
    ~
    \begin{subfigure}[b]{0.32\textwidth}
        \includegraphics[width=1\textwidth]{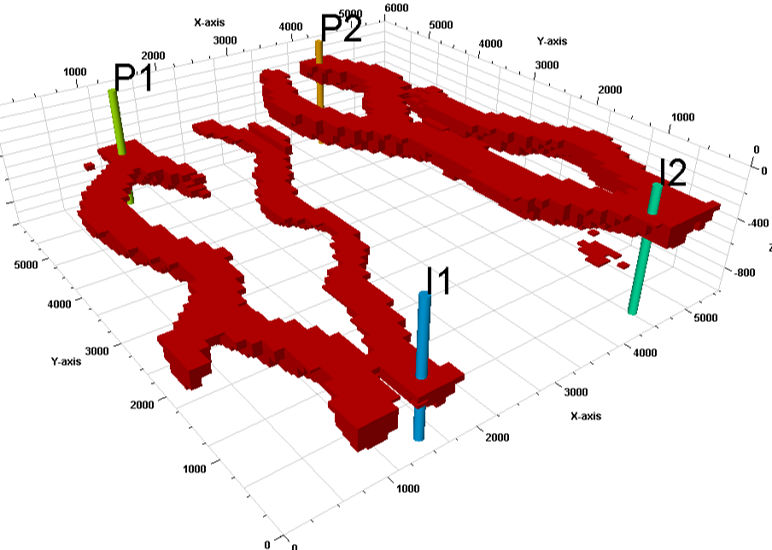}
        \caption{One CNN-PCA model}
    \end{subfigure}%
    \caption{3D channel geometry from (a) one new Petrel model, (b) one T-PCA model, and (c) corresponding CNN-PCA model. The T-PCA and CNN-PCA models correspond to the leftmost geomodels in Fig.~\ref{fig_case1_models}b and c (Case~1).}
    \label{fig_case1_chans}
\end{figure}

After training, 200 new Petrel realizations and 200 new PCA models are generated. The new PCA models are fed through the trained model transform net to obtain the CNN-PCA models. Truncation is performed on the new PCA models and on the CNN-PCA models to render them strictly binary. Specifically, cutoff values are determined for each set of models such that the final sand-facies fractions match that of the Petrel models (6.53\%). These models, along with the new Petrel realizations, comprise the test sets. 

Figure~\ref{fig_case1_models} presents four test-set PCA models (Fig.~\ref{fig_case1_models}a), the corresponding truncated-PCA models (denoted T-PCA, Fig.~\ref{fig_case1_models}b) and the corresponding CNN-PCA models (after truncation, Fig.~\ref{fig_case1_models}c). Figure~\ref{fig_case1_chans} displays the 3D channel geometry for a Petrel model (Fig.~\ref{fig_case1_chans}a), a truncated-PCA model (Fig.~\ref{fig_case1_chans}b), and the corresponding CNN-PCA model (Fig.~\ref{fig_case1_chans}~c). From Figs.~\ref{fig_case1_models} and \ref{fig_case1_chans}, it is apparent that the CNN-PCA models preserve geological realism much better than the truncated-PCA models. More specifically, the CNN-PCA models display intersecting channels of continuity, width, sinuosity and depth consistent with reference Petrel models.

\begin{figure}[!htb]
    \centering
    \includegraphics[width=0.32\textwidth]{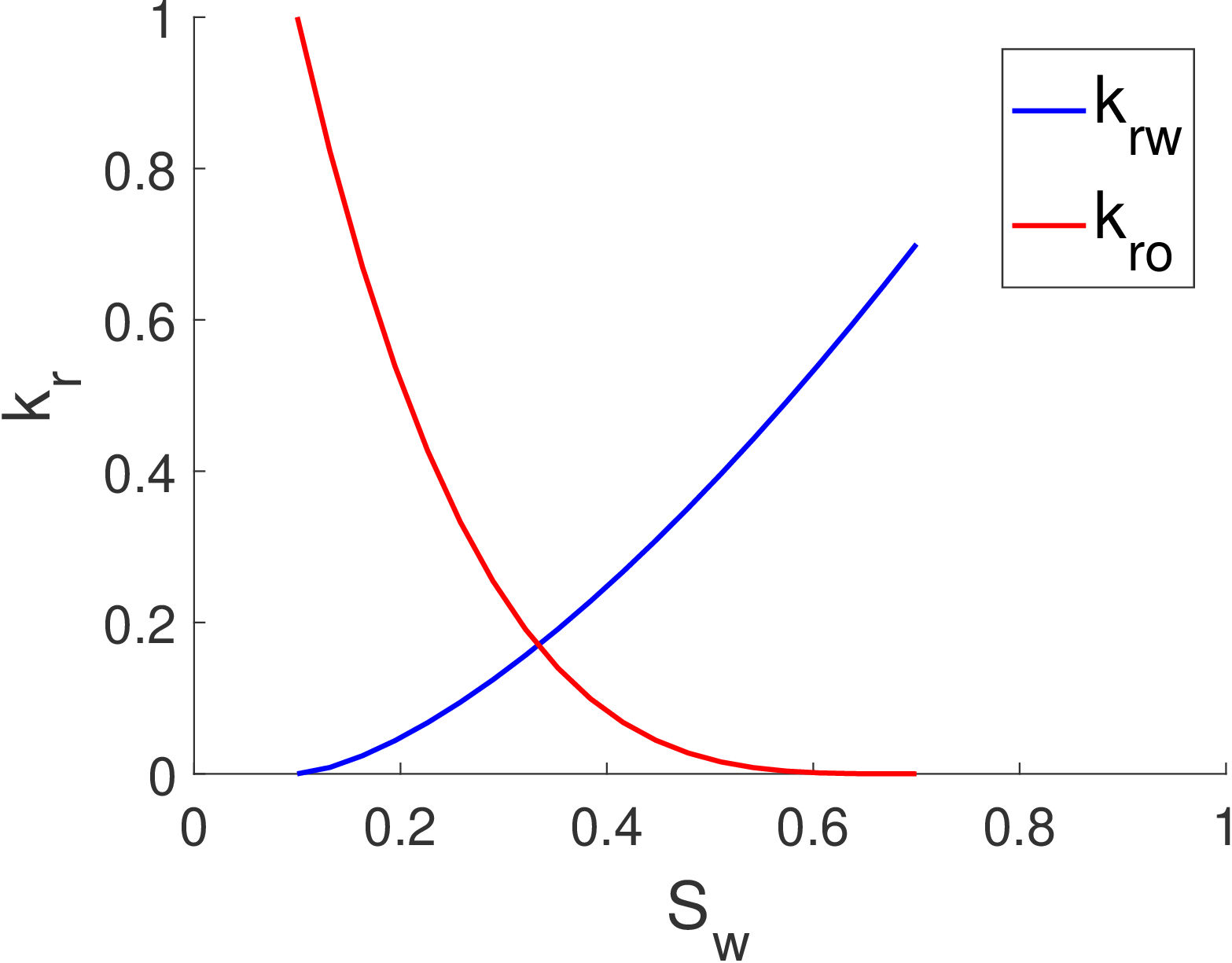}
    \caption{Relative permeability curves for all flow simulation models.}
    \label{fig_rel_perm}
\end{figure}

In addition to visual inspection, it is important to assess the CNN-PCA models quantitatively. We computed static channel connectivity metrics as well as flow responses for all of the test sets. The connectivity metrics suggested by \cite{Pardo-Iguzquiza2003} were found to be informative in our 2D study \citep{Liu2019}, but for the 3D cases considered here they appear to be too global to capture key interconnected-channel features that impact flow. Thus we focus on flow responses in the current assessment.

The flow set up involves aqueous and nonaqueous liquid phases. These can be viewed as NAPL and water in the context of an aquifer remediation project, or as oil and water in the context of oil production via water injection. Our terminology will correspond to the latter application.

Each grid block in the geomodel is of dimension 20~m in the $x$ and $y$ directions, and 5~m in the $z$ direction. Water viscosity is constant at 0.31~cp. Oil viscosity varies with pressure; it is 1.03~cp at a pressure of 325~bar. Relative permeability curves are shown in Fig.~\ref{fig_rel_perm}. The initial pressure of the reservoir (bottom layer) is 325~bar. The production and injection wells operate at constant bottom-hole pressures (BHPs) of 300~bar and 340~bar, respectively. The simulation time frame is 1500~days. The permeability and porosity for grid blocks in channels (sand) are $k=2000$~md and $\phi=0.2$. Grid blocks in mud are characterized by $k=20$~md and $\phi=0.15$. 

\begin{figure}[!htb]
    \begin{subfigure}[b]{1.0\textwidth}
    \centering
        \includegraphics[width=0.4\textwidth]{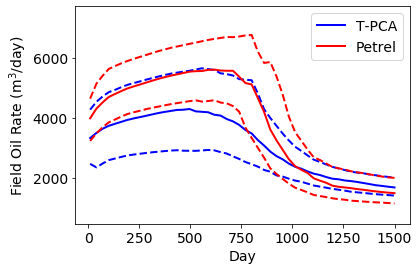}
        \includegraphics[width=0.4\textwidth]{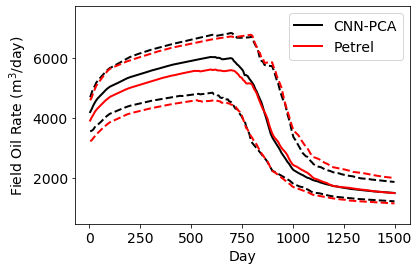}
        \caption{Field oil rate}
    \end{subfigure}%
    
    \begin{subfigure}[b]{1.0\textwidth}
    \centering
        \includegraphics[width=0.4\textwidth]{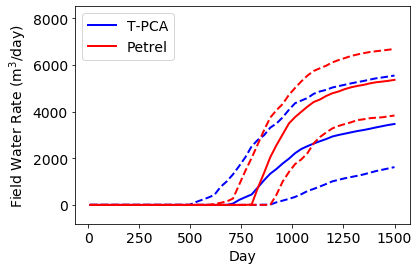}
        \includegraphics[width=0.4\textwidth]{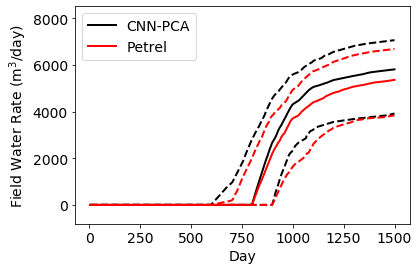}
        \caption{Field water rate}
    \end{subfigure}%

    \begin{subfigure}[b]{1.0\textwidth}
    \centering
        \includegraphics[width=0.4\textwidth]{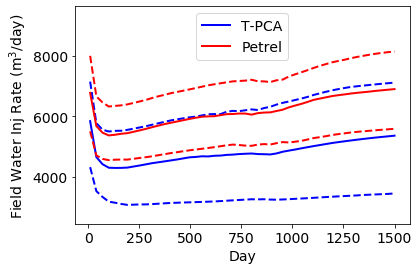}
        \includegraphics[width=0.4\textwidth]{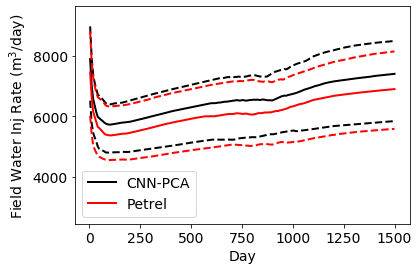}
        \caption{Field water injection rate}
    \end{subfigure}%

    \caption{Comparison of Petrel and T-PCA (left) and Petrel and CNN-PCA (right) field-wide flow statistics over ensembles of 200 (new) test cases. Red, blue and black curves represent results from Petrel, T-PCA, and CNN-PCA, respectively. Solid curves correspond to $\text{P}_{50}$ results, lower and upper dashed curves to $\text{P}_{10}$ and $\text{P}_{90}$ results (Case~1).}
    \label{fig_case1_flow_stats_field}
\end{figure}

\begin{figure}[!htb]
    \centering
    \begin{subfigure}[b]{0.32\textwidth}
        \includegraphics[width=1\textwidth]{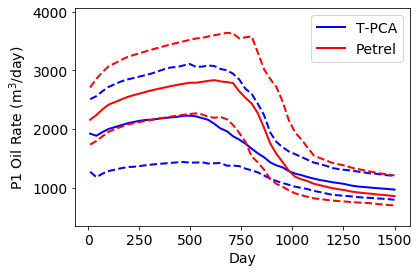}
    \end{subfigure}%
    ~ 
    \begin{subfigure}[b]{0.32\textwidth}
        \includegraphics[width=1\textwidth]{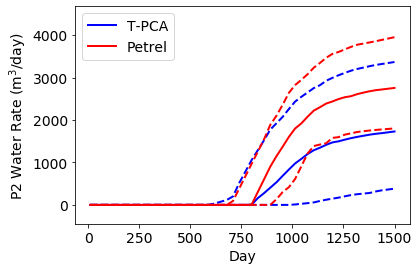}
    \end{subfigure}%
    ~ 
    \begin{subfigure}[b]{0.32\textwidth}
        \includegraphics[width=1\textwidth]{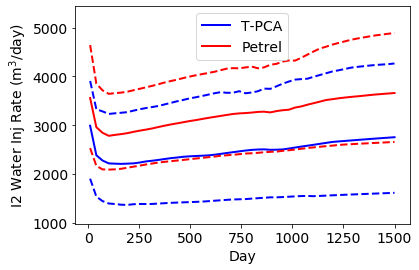}
    \end{subfigure}%
    
    \centering
    \begin{subfigure}[b]{0.32\textwidth}
        \includegraphics[width=1\textwidth]{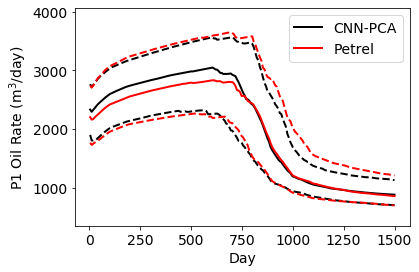}
        \caption{P1 oil production rate}
    \end{subfigure}%
    ~ 
    \begin{subfigure}[b]{0.32\textwidth}
        \includegraphics[width=1\textwidth]{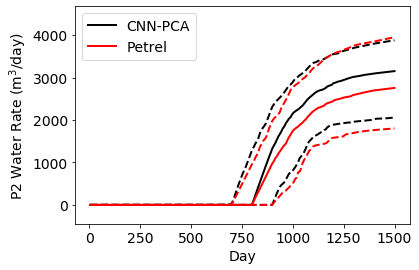}
        \caption{P2 water production rate}
    \end{subfigure}%
    ~ 
    \begin{subfigure}[b]{0.32\textwidth}
        \includegraphics[width=1\textwidth]{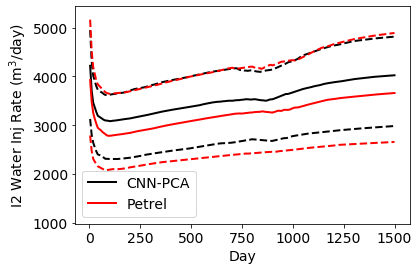}
        \caption{I2 water injection rate}
    \end{subfigure}%
    
    \caption{Comparison of Petrel and T-PCA (top) and Petrel and CNN-PCA (bottom) well-by-well flow statistics over ensembles of 200 (new) test cases. Red, blue and black curves represent results from Petrel, T-PCA, and CNN-PCA, respectively. Solid curves correspond to $\text{P}_{50}$ results, lower and upper dashed curves to $\text{P}_{10}$ and $\text{P}_{90}$ results (Case~1).}
    \label{fig_case1_flow_stats_well}
\end{figure}

Flow results are presented in terms of $\text{P}_{10}$, $\text{P}_{50}$ and $\text{P}_{90}$ percentiles for each test set. These results are determined, at each time step, based on the results for all 200 models. Results at different times correspond, in general, to different geomodels within the test set.

Figure~\ref{fig_case1_flow_stats_field} displays field-level results for the 200 test-set Petrel models (red curves), truncated-PCA models (T-PCA, blue curves), and CNN-PCA models (black curves). Figure~\ref{fig_case1_flow_stats_well} presents individual well responses (the wells shown have the highest cumulative phase production or injection). The significant visual discrepancies between the truncated-PCA and Petrel geomodels, observed in Figs.~\ref{fig_case1_models} and \ref{fig_case1_chans}, are reflected in the flow responses, where large deviations are evident. The CNN-PCA models, by contrast, provide flow results in close agreement with the Petrel models, for both field and well-level predictions. The CNN-PCA models display slightly higher field water rates, which may be due to a minor overestimation of channel connectivity or width. The overall match in $\text{P}_{10}$--$\text{P}_{90}$ results, however, demonstrates that the CNN-PCA geomodels exhibit the same level of variability as the Petrel models. This feature is important for the proper quantification of uncertainty. 


\subsection{Impact of Style Loss on CNN-PCA Geomodels}

\begin{figure}[!htb]
    \centering

    \begin{subfigure}[b]{0.65\textwidth}
        \includegraphics[width=1\textwidth]{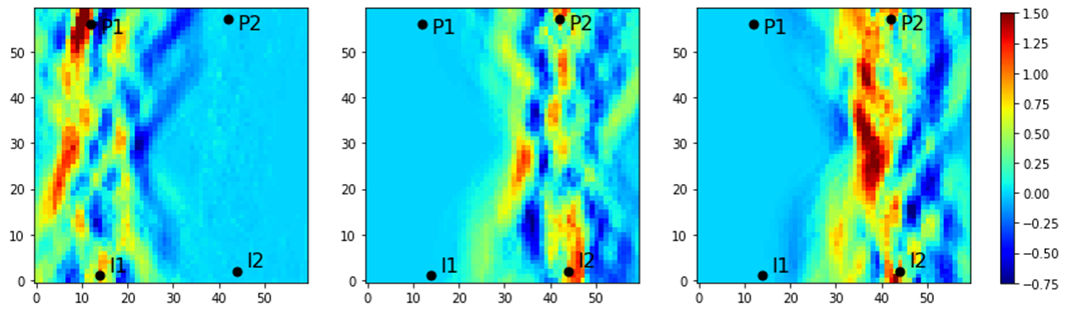}
        \caption{Areal maps from PCA models}
    \end{subfigure}%

    \begin{subfigure}[b]{0.65\textwidth}
        \includegraphics[width=1\textwidth]{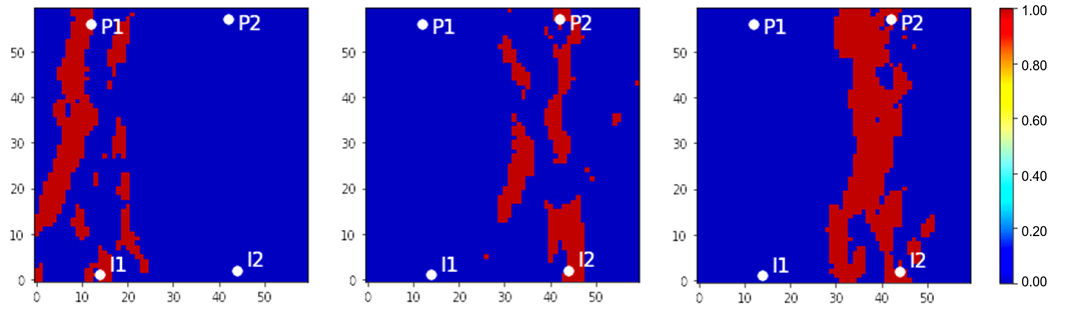}
        \caption{Areal maps from T-PCA models}
    \end{subfigure}%

    \begin{subfigure}[b]{0.65\textwidth}
        \includegraphics[width=1\textwidth]{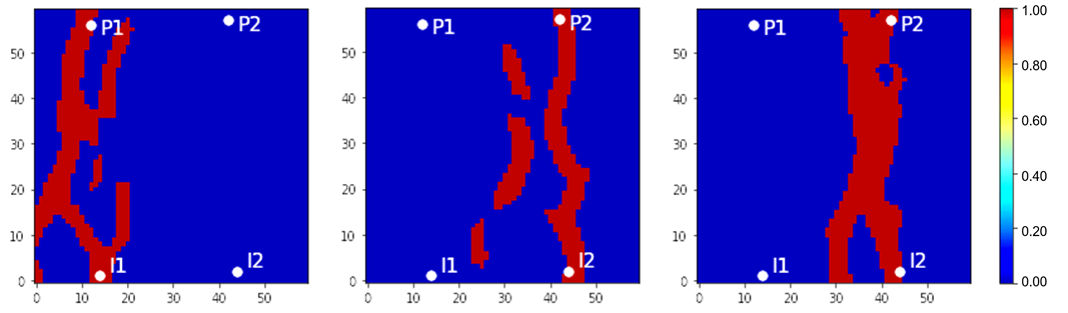}
        \caption{Areal maps from CNN-PCA models with $\gamma_s=0$}
    \end{subfigure}%

    \begin{subfigure}[b]{0.65\textwidth}
        \includegraphics[width=1\textwidth]{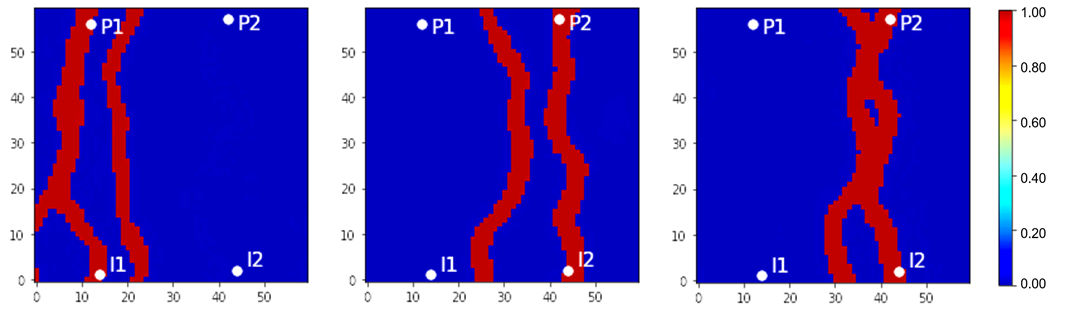}
        \caption{Areal maps from CNN-PCA models with $\gamma_s=100$}
    \end{subfigure}%
    \caption{Areal maps for layer 19 (left column) and layer 1 (middle and right columns) from (a) three different PCA models, (b) corresponding T-PCA models, (c) corresponding CNN-PCA models without style loss and (d) corresponding CNN-PCA models with style loss (Case~1).}
    \label{fig_model_style_loss_impact}
\end{figure}

\begin{figure}[!htb]
    \centering
    \begin{subfigure}[b]{0.32\textwidth}
        \includegraphics[width=1\textwidth]{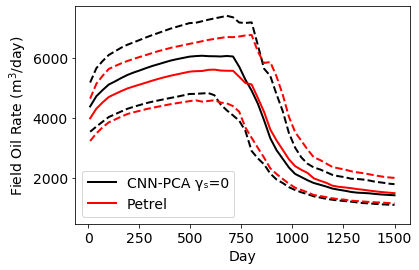}
        \caption{Field oil rate}
    \end{subfigure}%
    ~ 
    \begin{subfigure}[b]{0.32\textwidth}
        \includegraphics[width=1\textwidth]{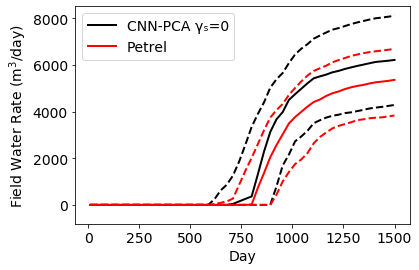}
        \caption{Field water rate}
    \end{subfigure}%
    ~ 
    \begin{subfigure}[b]{0.32\textwidth}
        \includegraphics[width=1\textwidth]{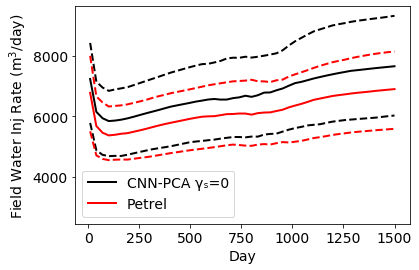}
        \caption{Field water injection rate}
    \end{subfigure}%
    
    \caption{Comparison of Petrel (red curves) and CNN-PCA without style loss (black curves) field-wide flow statistics over ensembles of 200 (new) test cases. Solid curves correspond to $\text{P}_{50}$ results, lower and upper dashed curves to $\text{P}_{10}$ and $\text{P}_{90}$ results (Case~1).}
    \label{fig_case1_flow_stats_no_style_loss}
\end{figure}

We now briefly illustrate the impact of style loss by comparing CNN-PCA geomodels generated with and without this loss term. Comparisons of areal maps for different layers in different models are shown in Fig.~\ref{fig_model_style_loss_impact}. Maps for three PCA models appear in Fig.~\ref{fig_model_style_loss_impact}a, the corresponding T-PCA models in Fig.~\ref{fig_model_style_loss_impact}b, CNN-PCA models without style loss ($\gamma_s=0$) in Fig.~\ref{fig_model_style_loss_impact}c, and CNN-PCA models with style loss ($\gamma_s=100$) in Fig.~\ref{fig_model_style_loss_impact}d. The flow statistics for CNN-PCA models without style loss are presented in Fig.~\ref{fig_case1_flow_stats_no_style_loss}. Hard data loss is included in all cases. The CNN-PCA models with reconstruction loss alone ($\gamma_s=0$) are more realistic visually, and result in more accurate flow responses, than the truncated-PCA models. However, the inclusion of style loss clearly acts to improve the CNN-PCA geomodels. This is evident in both the areal maps and the field-level flow responses (compare Fig.~\ref{fig_case1_flow_stats_no_style_loss} to Fig.~\ref{fig_case1_flow_stats_field} (right)).

We note finally that in a smaller 3D example considered in \cite{TangLiu2020}, the use of reconstruction loss alone was sufficient to achieve well-defined channels in the CNN-PCA geomodels (and accurate flow statistics). That case involved six wells (and thus more hard data) spaced closer together than in the current example. This suggests that style loss may be most important when hard data are limited, or do not act to strongly constrain channel geometry.




\subsection{Case 2 -- Three-Facies Channel-Levee-Mud System}

\begin{figure}[!htb]
    \centering
    \begin{subfigure}[b]{0.24\textwidth}
        \includegraphics[width=1\textwidth]{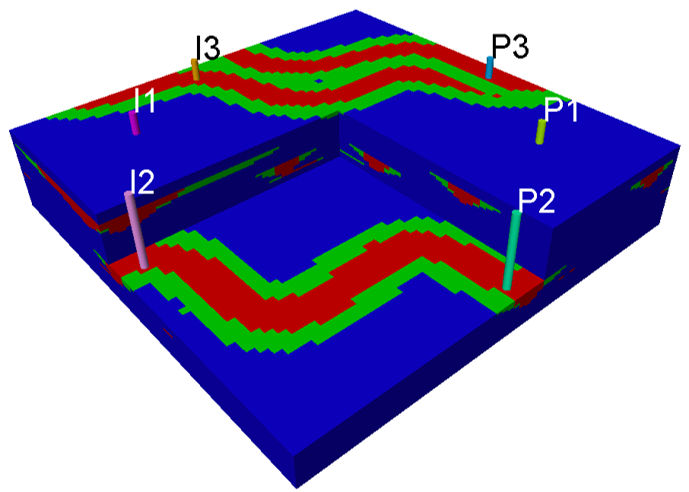}
    \end{subfigure}%
    ~ 
    \begin{subfigure}[b]{0.24\textwidth}
        \includegraphics[width=1\textwidth]{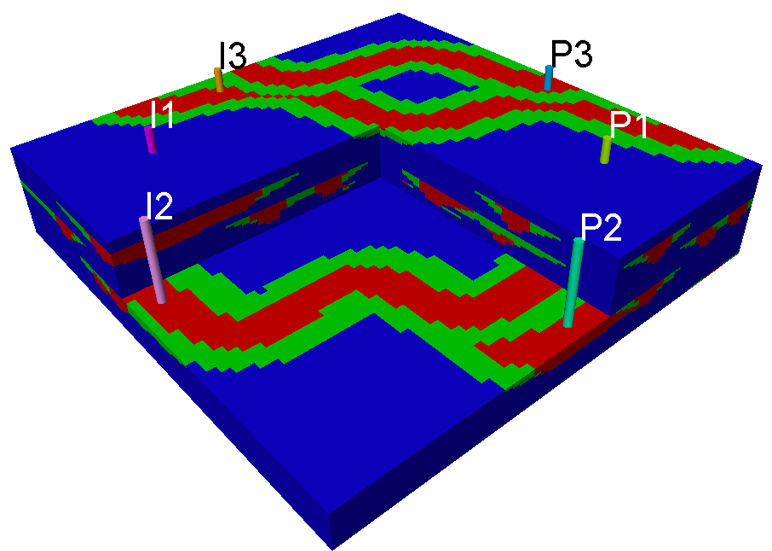}
    \end{subfigure}%
    ~
    \begin{subfigure}[b]{0.24\textwidth}
        \includegraphics[width=1\textwidth]{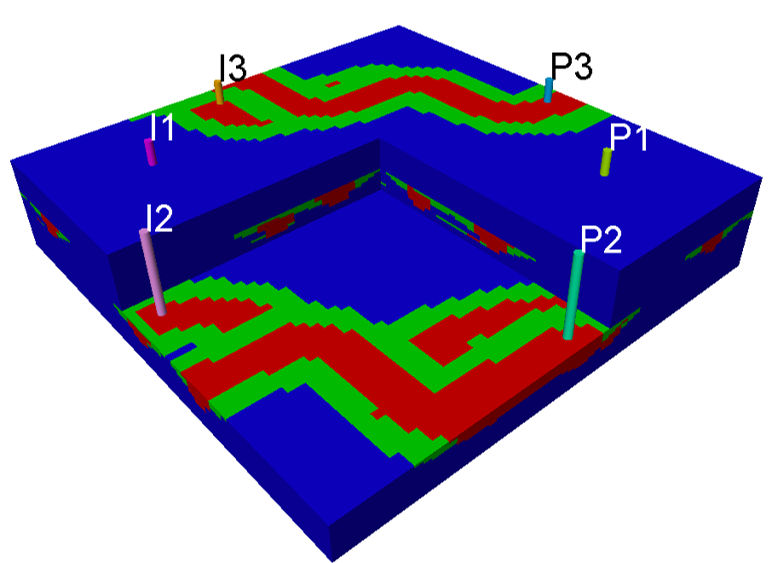}
    \end{subfigure}%
    ~
    \begin{subfigure}[b]{0.24\textwidth}
        \includegraphics[width=1\textwidth]{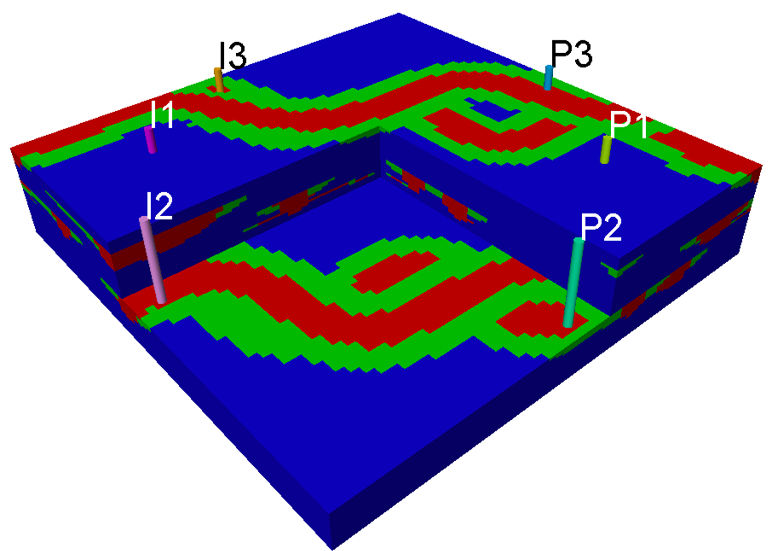}
    \end{subfigure}%

    \caption{Four Petrel realizations of the three-facies system, with sand shown in red, levee in green, and mud in blue (Case~2).}
    \label{fig_case2_petrel_models}
\end{figure}

This case involves three rock types, with the channel and levee facies displaying complex interconnected and overlapping geometries. The fluvial channel facies is of the highest permeability. The upper portion of each channel is surrounded by a levee facies, which is of intermediate permeability. The low-permeability mud facies comprises the remainder of the system. The average volume fractions for the channel and levee facies are 8.70\% and 5.42\%, respectively. Figure~\ref{fig_case2_petrel_models} displays four realizations generated from Petrel (channel in red, levee in green, mud in blue). The width and thickness of the levees is $\sim$55--60\% of the channel width and thickness. These models again contain $60 \times 60 \times 40$ cells. In this case there are three injection and three production wells. Well locations and hard data are summarized in Table~\ref{tab_well_case2}.

We considered two different ways of encoding mud, levee and channel facies. The `natural' approach is to encode mud as 0, levee as 1, and channel as 2. This treatment, however, has some disadvantages. Before truncation, CNN-PCA models are not strictly discrete, and a transition zone exists between mud and channel. This transition region will be interpreted as levee after truncation, which in turn leads to levee surrounding channels on all sides. This facies arrangement deviates from the underlying geology, where levees only appear near the upper portion of channels. In addition, since we preserve levee volume fraction, the average levee width becomes significantly smaller than it should be. For these reasons we adopt an alternative encoding strategy. This entails representing mud as 0, levee as 2, and channel as 1. This leads to better preservation of the location and geometry of levees.

\begin{table}[!htb]
\centering
\begin{tabular}{ c | c | c | c |c |c|c }
  & P1 & P2 & P3 & I1 & I2 & I3\\
  \hline
  Areal location ($x,~y$)&(48, 48)&(58, 31)&(32, 57)&(12, 12)&(28, 3) & (4, 26)\\
  Perforated layers in channel&15 - 21&25 - 30&1 - 5&15 - 20 & 25 - 30 & 1 - 6\\
  Perforated layers in levee & - & - & 21 - 24 & - & 6 - 8 & 36 - 40 \\
\end{tabular}

\caption{Well locations ($x$ and $y$ refer to areal grid-block indices) and perforations (Case~2)}
\label{tab_well_case2}
\end{table}

We again generate $\Nr=3000$ Petrel models, reconstructed PCA models and a separate set of new PCA models for training. We retain $l=800$ leading singular values, which capture $\sim$80\% of the total energy. The first 70 of these explain $\sim$40\% of the energy, so perturbation is performed on $\xi_j$, $j=71,...,800$ when generating the reconstructed PCA models. The training loss weighting factors in this case are $\gamma_\text{rec}=500$, $\gamma_s=50$ and $\gamma_h=10$. Other training parameters are the same as in Case~1. 

\begin{figure}[!htb]
    \centering
    

    \includegraphics[width=0.24\textwidth]{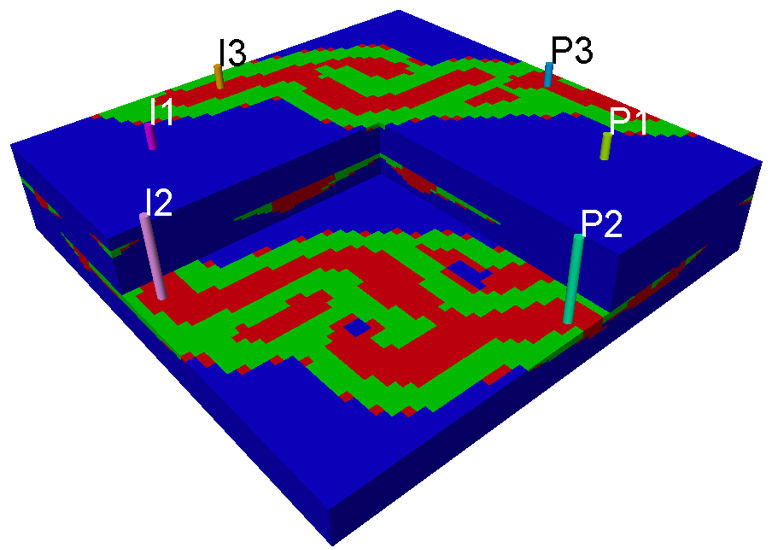}
    \includegraphics[width=0.24\textwidth]{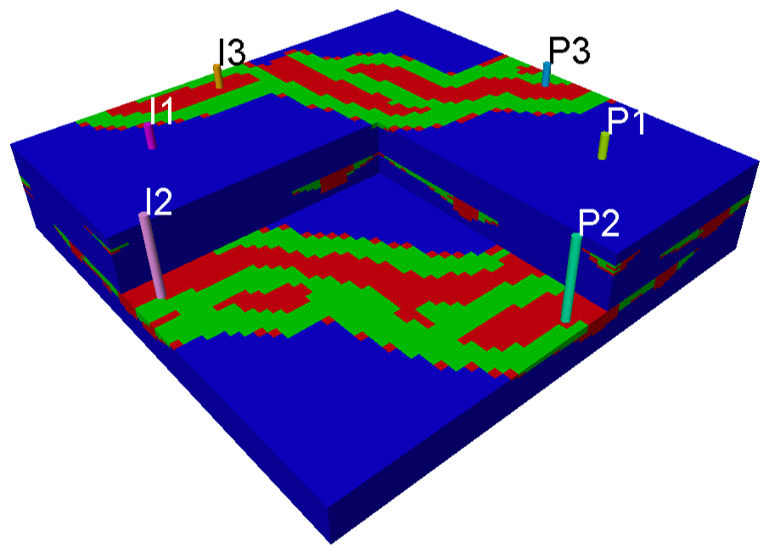}
    \includegraphics[width=0.24\textwidth]{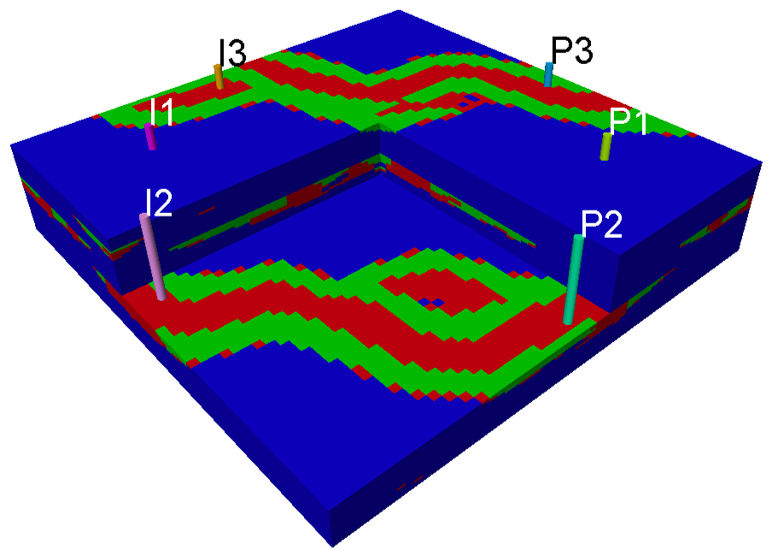}
    \includegraphics[width=0.24\textwidth]{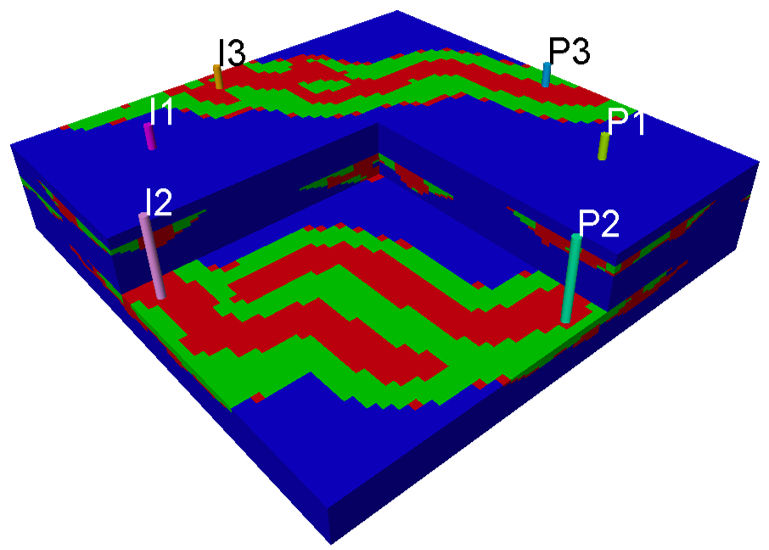}

    \caption{Four test-set CNN-PCA realizations of the three-facies system, with sand shown in red, levee in green, and mud in blue (Case~2).}
    \label{fig_models_case2}
\end{figure}

Test sets of 200 new realizations are then generated. The test-set PCA models are post-processed with the model transform net to obtain the CNN-PCA test set. The CNN-PCA geomodels are then truncated to be strictly ternary, with cutoff values determined such that the average facies fractions match those from the Petrel models. Four test-set CNN-PCA realizations are shown in Fig.~\ref{fig_models_case2}. These geomodels contain features consistent with those in the Petrel models (Fig.~\ref{fig_case2_petrel_models}). In addition to channel geometry, the CNN-PCA models also capture the geometry of the levees and their location relative to channels. 

\begin{figure}[!htb]
    \centering
    \begin{subfigure}[b]{0.32\textwidth}
        \includegraphics[width=1\textwidth]{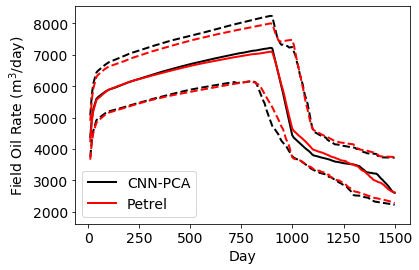}
        \caption{Field oil rate}
    \end{subfigure}%
    ~ 
    \begin{subfigure}[b]{0.32\textwidth}
        \includegraphics[width=1\textwidth]{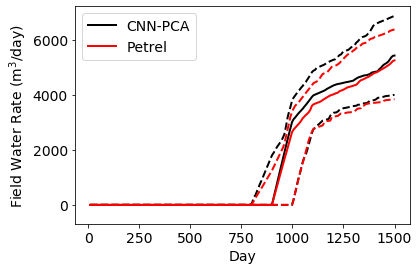}
        \caption{Field water rate}
    \end{subfigure}%
    ~ 
    \begin{subfigure}[b]{0.32\textwidth}
        \includegraphics[width=1\textwidth]{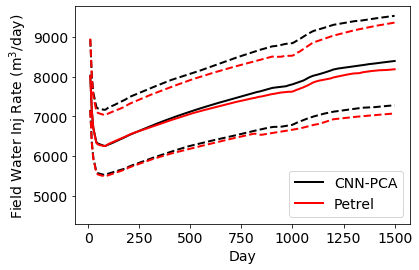}
        \caption{Field water injection rate}
    \end{subfigure}%
    
    \centering
    \begin{subfigure}[b]{0.32\textwidth}
        \includegraphics[width=1\textwidth]{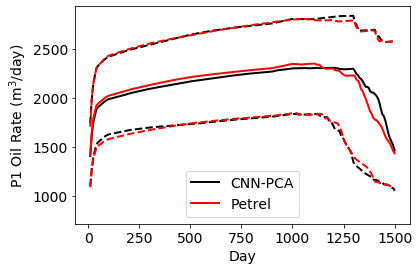}
        \caption{P1 oil rate}
    \end{subfigure}%
    ~ 
    \begin{subfigure}[b]{0.32\textwidth}
        \includegraphics[width=1\textwidth]{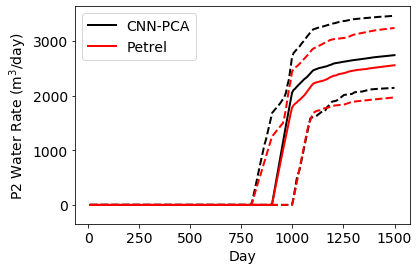}
        \caption{P2 water rate}
    \end{subfigure}%
    ~ 
    \begin{subfigure}[b]{0.32\textwidth}
        \includegraphics[width=1\textwidth]{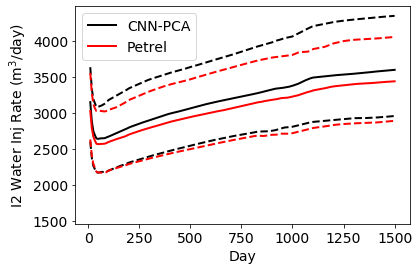}
        \caption{I2 water injection rate}
    \end{subfigure}%
    
    \caption{Comparison of Petrel (red curves) and CNN-PCA (black curves) flow statistics over ensembles of 200 (new) test cases. Solid curves correspond to $\text{P}_{50}$ results, lower and upper dashed curves to $\text{P}_{10}$ and $\text{P}_{90}$ results (Case~2).}
    \label{fig_case2_flow_stats}
\end{figure}

We now present flow results for this case. Permeability for channel, levee and mud are specified as 2000~md, 200~md and 20~md, respectively. Corresponding porosity values are 0.25, 0.15 and 0.05. Other simulation specifications are as in Case~1. Field-wide and individual well flow statistics (for wells with the largest cumulative phase production/injection) are presented in Fig.~\ref{fig_case2_flow_stats}. Consistent with the results for Case~1, we observe generally close agreement between flow predictions for CNN-PCA and Petrel models. Again, the close matches in the $\text{P}_{10}$--$\text{P}_{90}$ ranges indicate that the CNN-PCA geomodels capture the inherent variability of the Petrel models. 

Truncated-PCA geomodels, and comparisons between flow results for these models against Petrel models, are shown in Figs.~S1 and~S2 in SI. The truncated-PCA models appear less geologically realistic, and provide much less accurate flow predictions, than the CNN-PCA geomodels. 


    



\subsection{Case 3 -- Bimodal Channelized System}
\label{sec_case3}

In the previous two cases, permeability and porosity within each facies were constant. We now consider a bimodal system, where log-permeability and porosity within the two facies follow Gaussian distributions. The facies model, well locations and perforations are the same as in Case~1. The log-permeability in the sand facies follows a Gaussian distribution with mean 6.7 and variance 0.2, while in the mud facies the mean is 3.5 and the variance is 0.19. The log-permeability at well locations in all layers is treated as hard data. 

\begin{figure}[!htb]
    \centering
    \begin{subfigure}[b]{0.24\textwidth}
        \includegraphics[width=1\textwidth]{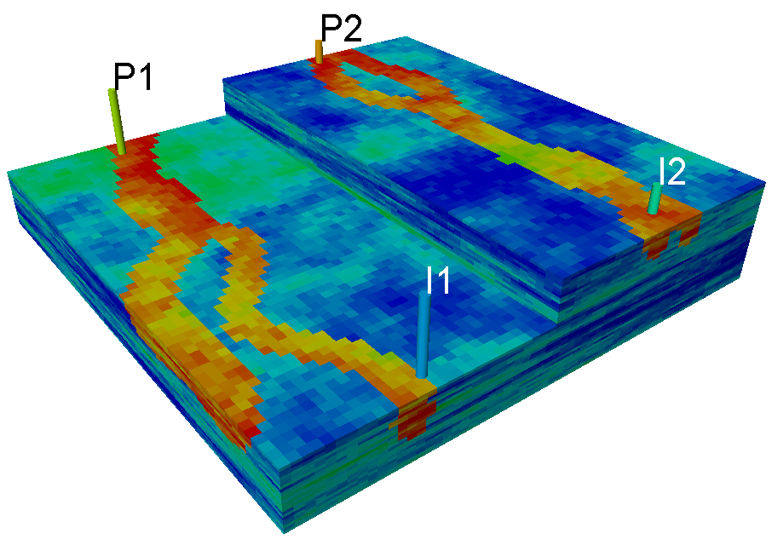}
    \end{subfigure}%
    ~ 
    \begin{subfigure}[b]{0.24\textwidth}
        \includegraphics[width=1\textwidth]{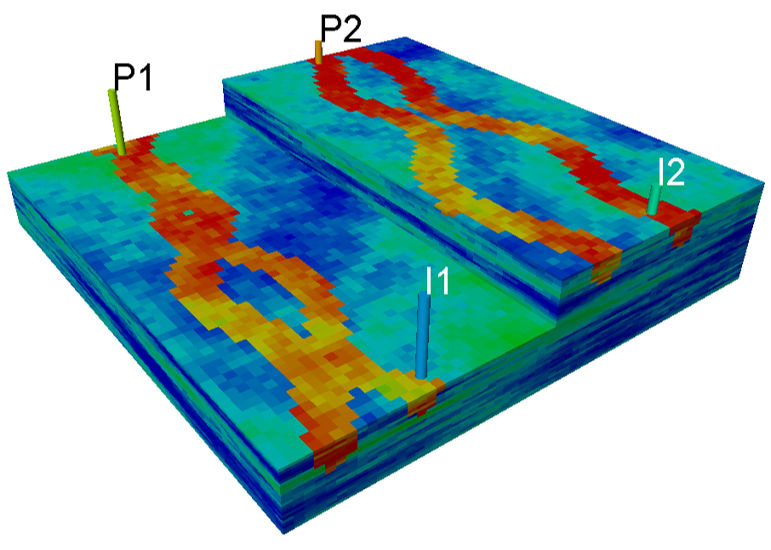}
    \end{subfigure}%
    ~
    \begin{subfigure}[b]{0.24\textwidth}
        \includegraphics[width=1\textwidth]{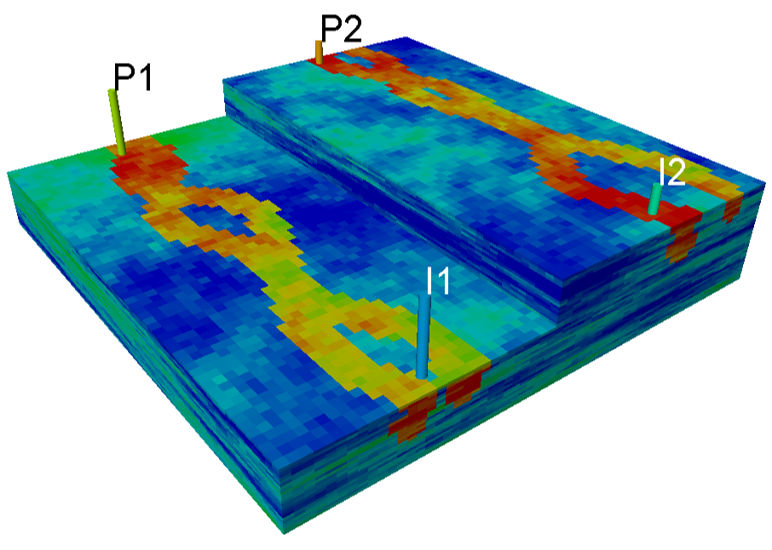}
    \end{subfigure}%
    ~
    \begin{subfigure}[b]{0.24\textwidth}
        \includegraphics[width=1\textwidth]{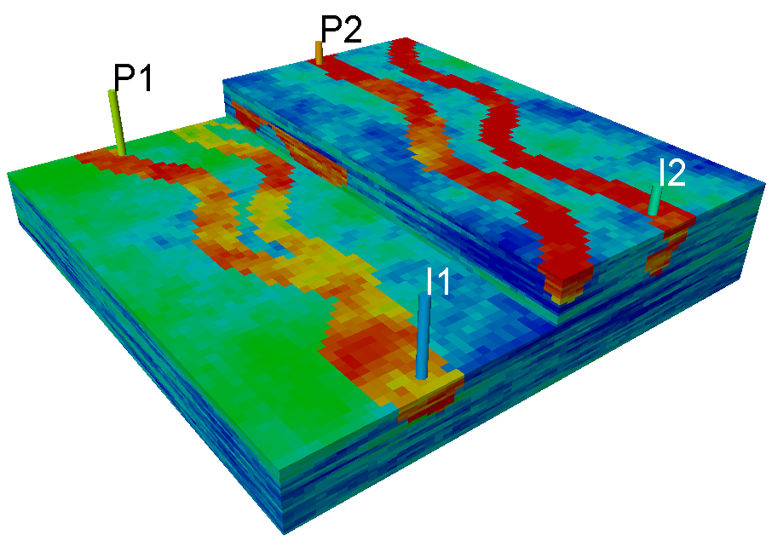}
    \end{subfigure}%
    \vspace{0.2cm}
    \begin{subfigure}[b]{0.6\textwidth}
        \includegraphics[width=1\textwidth]{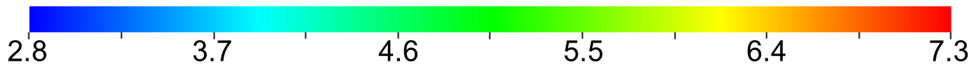}
    \end{subfigure}%

    \caption{Four Petrel log-permeability realizations of the bimodal channelized system (Case~3).}
    \label{fig_case3_petrel}
\end{figure}

We use the sequential Gaussian simulation algorithm in Petrel to generate log-permeability values within each facies. The final log-permeability field is obtained using a cookie-cutter approach. We use $\Bm^\text{f} \in \mathbb{R}^{\Nc}$ to denote the facies model, and $\Bm^\text{s} \in \mathbb{R}^{\Nc}$ and $\Bm^\text{m} \in \mathbb{R}^{\Nc}$ to denote log-permeability within the sand and mud facies. The log-permeability of grid block $i$, in the bimodal system $\Bm \in \mathbb{R}^{\Nc}$, is then
\begin{equation}
    (m_\text{gm})_i = (m_\text{gm}^\text{f})_i (m_\text{gm}^\text{s})_i + [1 - (m_\text{gm}^\text{f})_i] (m_\text{gm}^\text{m})_i, \hp i=1,...,\Nc.
    \label{eq_bimodal_cc}
\end{equation}
Figure~\ref{fig_case3_petrel} shows four log-permeability realizations generated by Petrel. Both channel geometry and within-facies heterogeneity are seen to vary between models.


For this bimodal system we apply a two-step approach. Specifically, CNN-PCA is used for facies parameterization, and two separate PCA models are used to parameterize log-permeability within the two facies. Since the facies model is the same as in Case~1, we use the same CNN-PCA model. Thus the reduced dimension for the facies model is $l^\text{f} = 400$. We then construct two PCA models to represent log-permeability in each facies. For these models we set $l^\text{s} = l^\text{m} = 200$, which explains \textapprox$60\%$ of the total energy. A smaller percentage is used here to limit the overall size of the low-dimensional variable $l$ (note that $l=l^\text{f}+l^\text{s}+l^\text{m}$), at the cost of discarding some amount of small-scale variation within each facies. This is expected to have a relatively minor impact on flow response.


To generate new PCA models, we sample $\Bxi^\text{f} \in \mathbb{R}^{l^\text{f}}$, $\Bxi^\text{s}\in \mathbb{R}^{l^\text{s}}$ and $\Bxi^\text{m}\in \mathbb{R}^{l^\text{m}}$ separately from standard normal distributions. We then apply Eq.~\ref{eq_pca} to construct PCA models $\Bmpca^\text{f} \in \mathbb{R}^{\Nc}$, $\Bmpca^\text{s} \in \mathbb{R}^{\Nc}$ and $\Bmpca^\text{m} \in \mathbb{R}^{\Nc}$. The model transform net then maps $\Bmpca^\text{f}$ to the CNN-PCA facies model (i.e., $\Bmcnnpca^\text{f} = f_W(\Bmpca^\text{f})$). After truncation, the cookie-cutter approach is applied to provide the final CNN-PCA bimodal log-permeability model:
\begin{equation}
    (m_\text{cnnpca})_i = (m_\text{cnnpca}^\text{f})_i (m_\text{pca}^\text{s})_i + [1 -(m_\text{cnnpca}^\text{f})_i] (m_\text{pca}^\text{m})_i, \hp i=1,...,\Nc.
    \label{eq_bimodal_cc}
\end{equation}

\begin{figure}[!htb]
    \centering
    \begin{subfigure}[b]{0.24\textwidth}
        \includegraphics[width=1\textwidth]{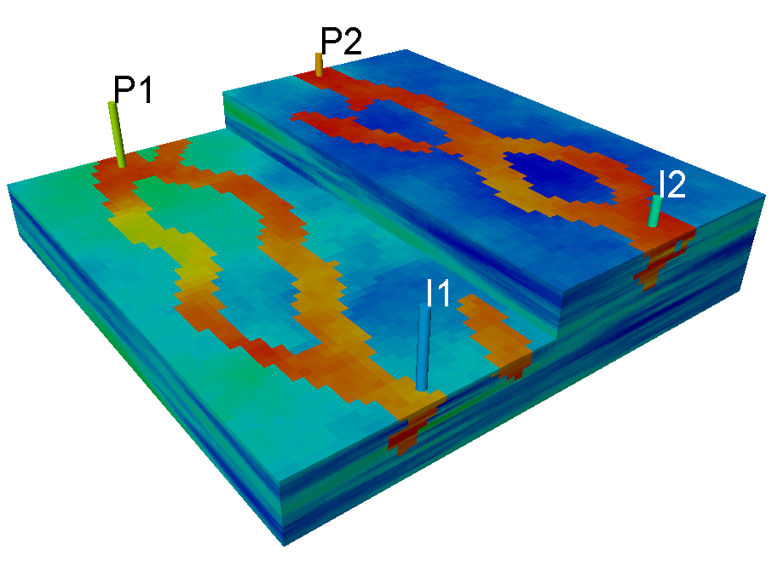}
    \end{subfigure}%
    ~ 
    \begin{subfigure}[b]{0.24\textwidth}
        \includegraphics[width=1\textwidth]{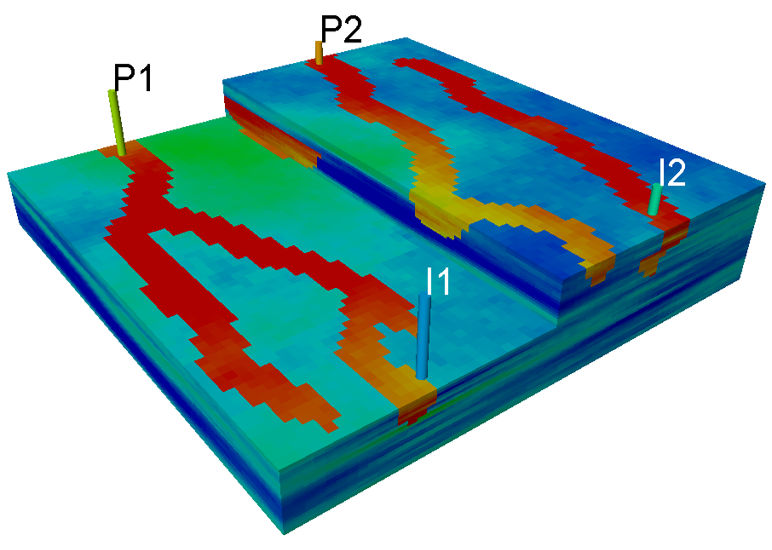}
    \end{subfigure}%
    ~
    \begin{subfigure}[b]{0.24\textwidth}
        \includegraphics[width=1\textwidth]{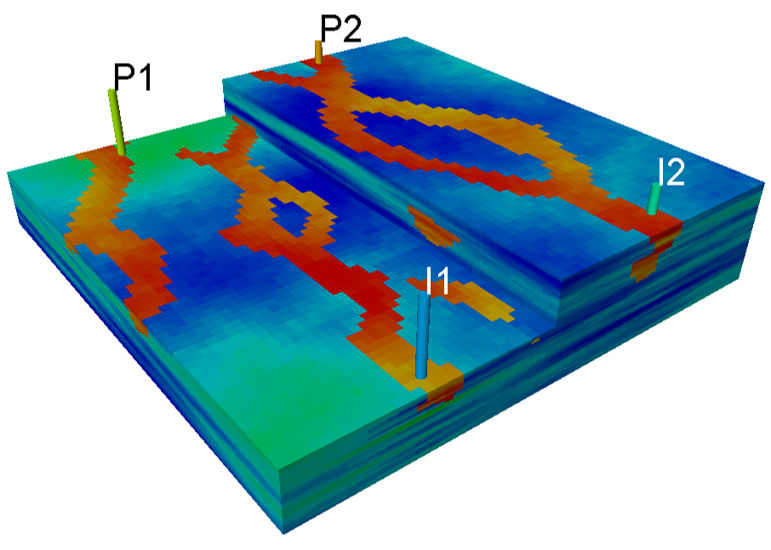}
    \end{subfigure}%
    ~
    \begin{subfigure}[b]{0.24\textwidth}
        \includegraphics[width=1\textwidth]{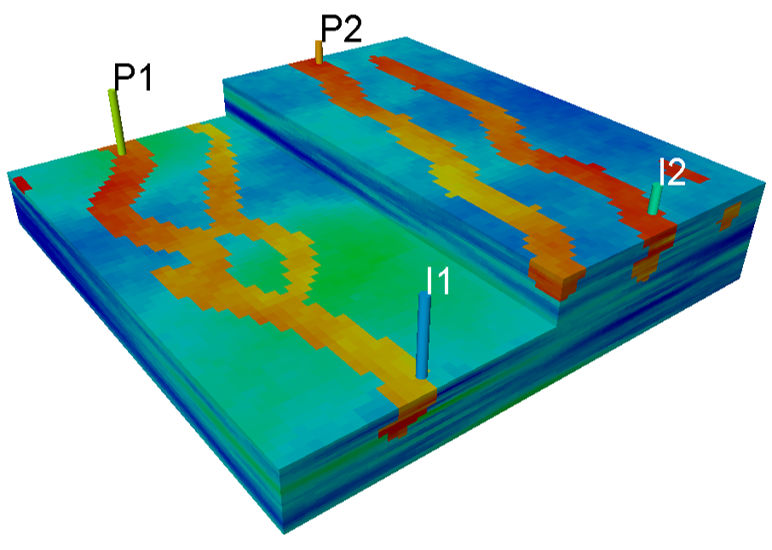}
    \end{subfigure}%
    \vspace{0.2cm}
    \begin{subfigure}[b]{0.6\textwidth}
        \includegraphics[width=1\textwidth]{colorbar.jpg}
    \end{subfigure}%

    \caption{Four test-set CNN-PCA log-permeability realizations of the bimodal channelized system (Case~3).}
    \label{fig_case3_cnnpca_models}
\end{figure}

\begin{figure}[!htb]
    \centering
    \begin{subfigure}[b]{0.45\textwidth}
        \includegraphics[width=1\textwidth]{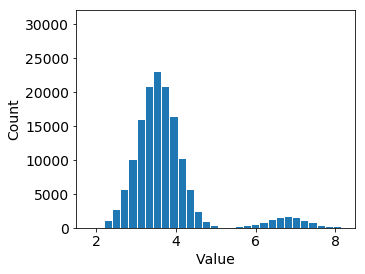}
        \caption{Petrel}
    \end{subfigure}%
    ~ 
    \begin{subfigure}[b]{0.45\textwidth}
        \includegraphics[width=1\textwidth]{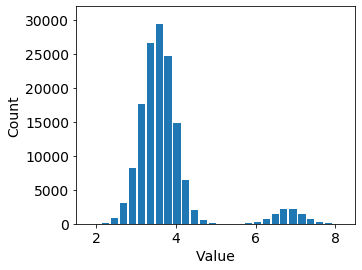}
        \caption{CNN-PCA}
    \end{subfigure}%
    
    \caption{Average histograms of the 200 test-set realizations from (a) Petrel  and (b) CNN-PCA (Case~3).}
    \label{fig_case3_histo}
\end{figure}

Figure~\ref{fig_case3_cnnpca_models} shows four of the resulting test-set CNN-PCA geomodels. Besides preserving channel geometry, the CNN-PCA models also display large-scale property variations within each facies. The CNN-PCA models are smoother than the reference Petrel models because small-scale variations in log-permeability within each facies are not captured in the PCA representations, as noted earlier. The average histograms for the 200 test-set Petrel and CNN-PCA geomodels are shown in Fig.~\ref{fig_case3_histo}. The CNN-PCA histogram is clearly bimodal, and in general correspondence with the Petrel histogram, though variance is underpredicted in the CNN-PCA models. This is again because variations at the smallest scales have been neglected.

To construct flow models, we assign permeability and porosity, for block $i$, as $k_i = \exp(m_i)$ and $\phi_i = m_i/40$. The median values for permeability within channel and mud facies are \textapprox800~md and \textapprox30~md, while those for porosity are \textapprox0.16 and \textapprox0.08. The simulation setup is otherwise the same as in Case~1.

\begin{figure}[!htb]
    \centering
    \begin{subfigure}[b]{0.32\textwidth}
        \includegraphics[width=1\textwidth]{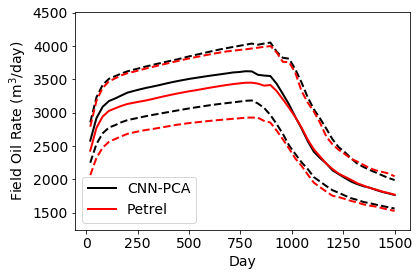}
        \caption{Field oil rate}
    \end{subfigure}%
    ~ 
    \begin{subfigure}[b]{0.32\textwidth}
        \includegraphics[width=1\textwidth]{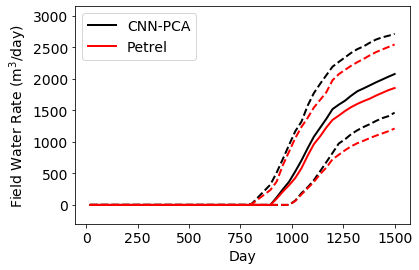}
        \caption{Field water rate}
    \end{subfigure}%
    ~ 
    \begin{subfigure}[b]{0.32\textwidth}
        \includegraphics[width=1\textwidth]{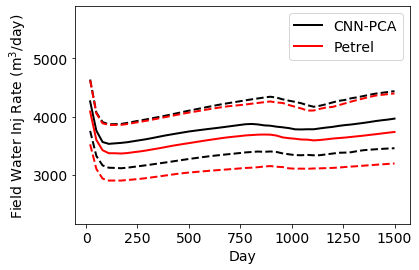}
        \caption{Field water injection rate}
    \end{subfigure}%
    
    \centering
    \begin{subfigure}[b]{0.32\textwidth}
        \includegraphics[width=1\textwidth]{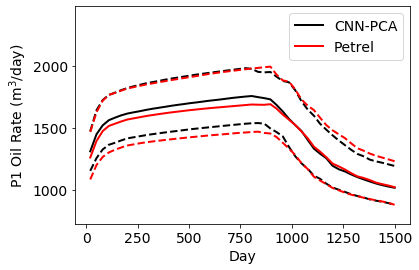}
        \caption{P1 oil rate}
    \end{subfigure}%
    ~ 
    \begin{subfigure}[b]{0.32\textwidth}
        \includegraphics[width=1\textwidth]{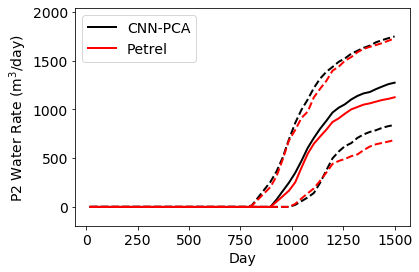}
        \caption{P2 water rate}
    \end{subfigure}%
    ~ 
    \begin{subfigure}[b]{0.32\textwidth}
        \includegraphics[width=1\textwidth]{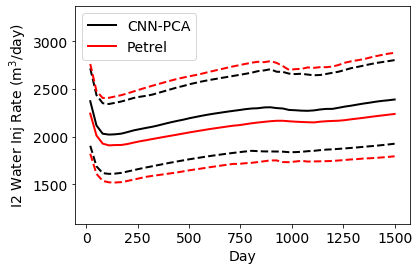}
        \caption{I2 water injection rate}
    \end{subfigure}%

    \caption{Comparison of Petrel (red curves) and CNN-PCA (black curves) flow statistics over ensembles of 200 (new) test cases. Solid curves correspond to $\text{P}_{50}$ results, lower and upper dashed curves to $\text{P}_{10}$ and $\text{P}_{90}$ results (Case~3).}
    \label{fig_case3_flow_stats}
\end{figure}

Flow statistics for Case~3 are shown in Fig.~\ref{fig_case3_flow_stats}. Consistent with the results for the previous cases, we observe close agreement between the $\text{P}_{10}$, $\text{P}_{50}$ and $\text{P}_{90}$ predictions from CNN-PCA and Petrel geomodels. There does appear to be a slight underestimation of variability in the CNN-PCA models, however, which may result from the lack of small-scale property variation. 

PCA geomodels and corresponding flow results for this case are shown in Figs.~S3--S5 in SI. Again, these models lack the realism and flow accuracy evident in the CNN-PCA geomodels.

\section{History Matching using CNN-PCA}
\label{sec-hm}

CNN-PCA is now applied for a history matching problem involving the bimodal channelized system. With the two-step CNN-PCA approach, the bimodal log-permeability and porosity fields are represented with three low-dimensional variables: $\Bxi^{\text{f}} \in \mathbb{R}^{400}$ for the facies model, and $\Bxi^{\text{s}} \in \mathbb{R}^{200}$ and $\Bxi^{\text{m}} \in \mathbb{R}^{200}$ for log-permeability and porosity in sand and mud. Concatenating the three low-dimensional variables gives $\Bxi_l = [\Bxi^{\text{f}},\Bxi^{\text{s}},\Bxi^{\text{m}}]  \in \mathbb{R}^{l}$, with $l=800$. This $\Bxi_l$ represents the uncertain variables considered during history matching. 

Observed data include oil and water production rates at the two producers, and water injection rate at the two injectors, collected every 100~days for the first 500~days. This gives a total of $\Nd=30$ observations. Standard deviations for error in observed data are 1\%, with a minimum value of 2~m$^3$/day. The leftmost Petrel realization in Fig.~\ref{fig_case3_petrel} is used as the true geomodel. Observed data are generated by performing flow simulation with this model and then perturbing the simulated production and injection data consistent with standard deviations of measurement errors.

\begin{figure}[!htb]
    \centering
    \begin{subfigure}[b]{0.44\textwidth}
        \includegraphics[width=1\textwidth]{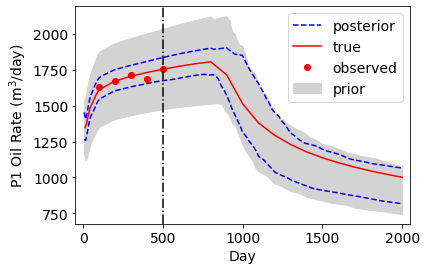}
        \caption{P1 oil rate}
    \end{subfigure}%
    ~ 
    \begin{subfigure}[b]{0.44\textwidth}
        \includegraphics[width=1\textwidth]{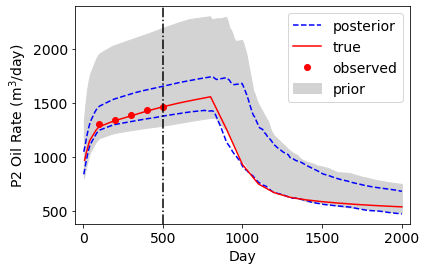}
        \caption{P2 oil rate}
    \end{subfigure}%
    
    \centering
    \begin{subfigure}[b]{0.44\textwidth}
        \includegraphics[width=1\textwidth]{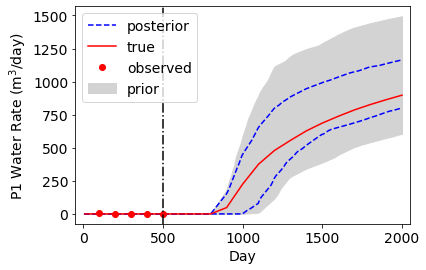}
        \caption{P1 water rate}
    \end{subfigure}%
    ~ 
    \begin{subfigure}[b]{0.44\textwidth}
        \includegraphics[width=1\textwidth]{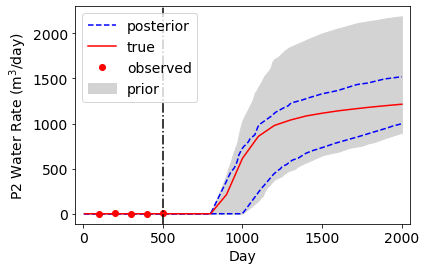}
        \caption{P2 water rate}
    \end{subfigure}%
    
    \centering
    \begin{subfigure}[b]{0.44\textwidth}
        \includegraphics[width=1\textwidth]{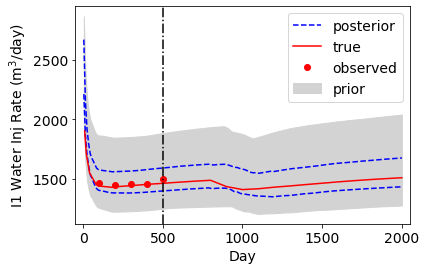}
        \caption{I1 water injection rate}
    \end{subfigure}%
    ~ 
    \begin{subfigure}[b]{0.44\textwidth}
        \includegraphics[width=1\textwidth]{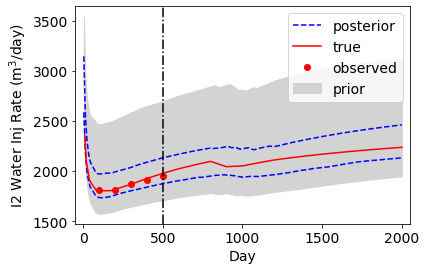}
        \caption{I2 water injection rate}
    \end{subfigure}%

    \caption{Prior and posterior flow results for bimodal channelized system. Gray regions represent the prior $\text{P}_{10}$-–$\text{P}_{90}$ range, red points and red curves denote observed and true data, and blue dashed curves denote the posterior $\text{P}_{10}$ (lower) and $\text{P}_{90}$ (upper) predictions. Vertical dashed line divides simulation time frame into history match and prediction periods.}
    \label{fig_hm_data}
\end{figure}

We use ESMDA \citep{Emerick2013} for history matching. This algorithm has been used previously by \cite{Canchumuni2017, Canchumuni2018, Canchumuni2019a, Canchumun2020} for data assimilation with deep-learning-based geological parameterizations. ESMDA is an ensemble-based procedure that starts with an ensemble of prior uncertain variables. At each data assimilation step, uncertain variables are updated by assimilating simulated production data to observed data with inflated measurement errors. We use an ensemble size of $N_\text{e} = 200$. The prior ensemble consists of 200 random realizations of $\Bxi_l \in \mathbb{R}^{800}$ sampled from the standard normal distribution. Each realization of $\Bxi_l$ is then divided into $\Bxi^{\text{f}}$, $\Bxi^{\text{s}}$ and $\Bxi^{\text{m}}$. CNN-PCA realizations of bimodal log-permeability and porosity are then generated and simulated.  

In ESMDA the ensemble is updated through application of
\begin{equation}
    \Bxi_l^{u,j} = \Bxi_l^j + C_{\xi d}(C_{dd} + \alpha C_{d})^{-1}(\Bd^j - \Bdobs^*), \hp j=1,...,N_\text{e},
\end{equation}
where $\Bd^j \in \mathbb{R}^{\Nd}$ represents simulated production data, $\Bdobs^*\in \mathbb{R}^{\Nd}$ denotes randomly perturbed observed data sampled from $N(\Bdobs, \alpha C_{d})$, where $C_{d}\in \mathbb{R}^{\Nd \times \Nd}$ is a diagonal prior covariance matrix for the measurement error and $\alpha$ is an error inflation factor. The matrix $C_{\xi d}\in \mathbb{R}^{l \times \Nd}$ is the cross-covariance between $\Bxi$ and $\Bd$ estimated by
\begin{equation}
    C_{\xi d} = \dfrac{1}{N_\text{e} - 1}\sum_{j=1}^{N_\text{e}}(\Bxi_l^j - \Bar{\Bxi}_l)(\Bd^j - \Bar{\Bd})^T,
    \label{eq_cxid}
\end{equation}
and $C_{d d} \in \mathbb{R}^{\Nd \times \Nd}$ is the auto-covariance of $\Bd$ estimated by
\begin{equation}
    C_{dd} = \dfrac{1}{N_\text{e} - 1}\sum_{j=1}^{N_\text{e}}(\Bd^j - \Bar{\Bd})(\Bd^j - \Bar{\Bd})^T.
    \label{eq_cdd}
\end{equation}
In Eqs.~\ref{eq_cxid} and \ref{eq_cdd}, the overbar denotes the mean over the $N_\text{e}$ samples at the current iteration. The updated variables $\Bxi_l^{u,j}$, $j=1,...,N_\text{e}$, then represent a new ensemble. The process is applied multiple times, each time with a different inflation factor $\alpha$ and a new random perturbation of the observed data. Here we assimilate data four times using inflation factors of 9.33, 7.0, 4.0 and 2.0, as suggested by \cite{Emerick2013}. 

Data assimilation results are shown in Fig.~\ref{fig_hm_data}. The gray region denotes the $\text{P}_{10}$--$\text{P}_{90}$ range for the prior models, while the dashed blue curves indicate the $\text{P}_{10}$--$\text{P}_{90}$ posterior range. Red points show the observed data, and the red curves the true model response. We observe uncertainty reduction in all quantities over at least a portion of the time frame, with the observed and true data consistently falling within the $\text{P}_{10}$--$\text{P}_{90}$ posterior range. Of particular interest is the fact that substantial uncertainty reduction is achieved in water-rate predictions even though none of the producers experiences water breakthrough in the history matching period.

\begin{figure}[!htb]
    \centering
    \begin{subfigure}[b]{0.32\textwidth}
        \includegraphics[width=1\textwidth]{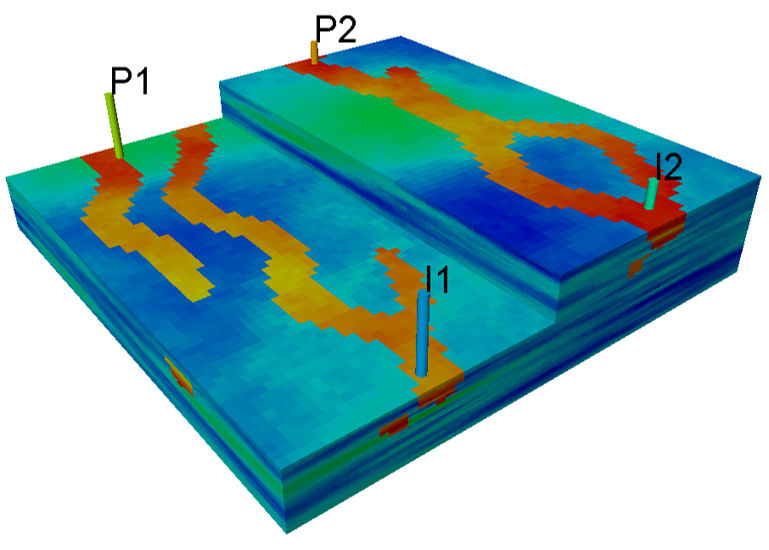}
        \caption{Prior model \#1}
    \end{subfigure}%
    ~ 
    \begin{subfigure}[b]{0.32\textwidth}
        \includegraphics[width=1\textwidth]{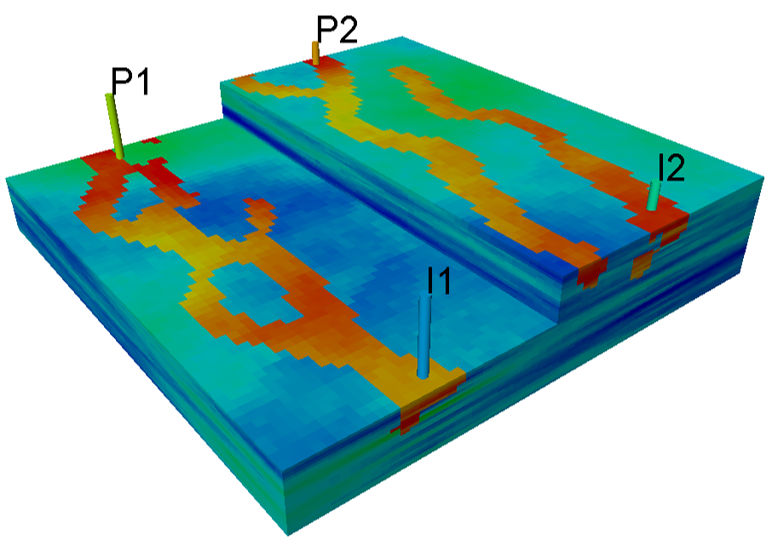}
        \caption{Prior model \#2}
    \end{subfigure}%
    ~ 
    \begin{subfigure}[b]{0.32\textwidth}
        \includegraphics[width=1\textwidth]{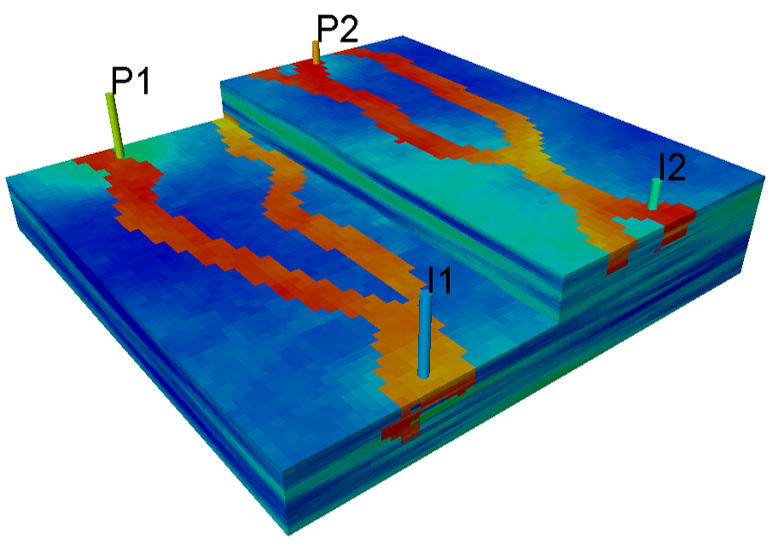}
        \caption{Prior model \#3}
    \end{subfigure}%

    \begin{subfigure}[b]{0.32\textwidth}
        \includegraphics[width=1\textwidth]{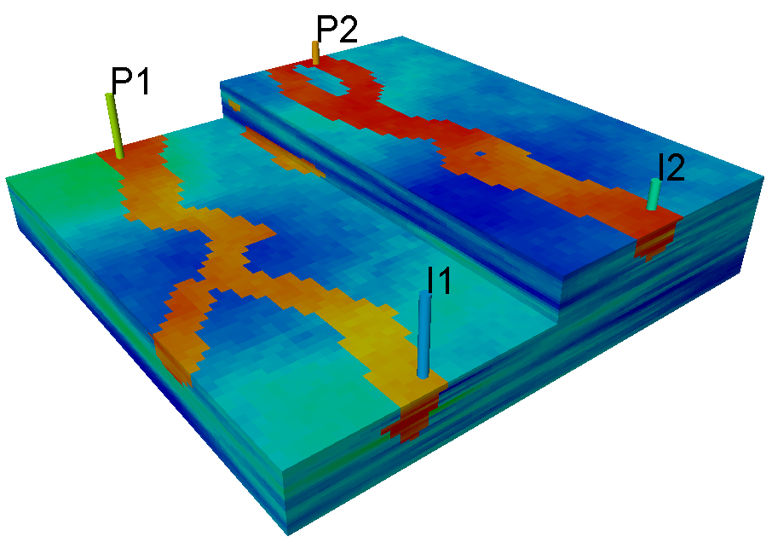}
        \caption{Posterior model \#1}
    \end{subfigure}%
    ~
    \begin{subfigure}[b]{0.32\textwidth}
        \includegraphics[width=1\textwidth]{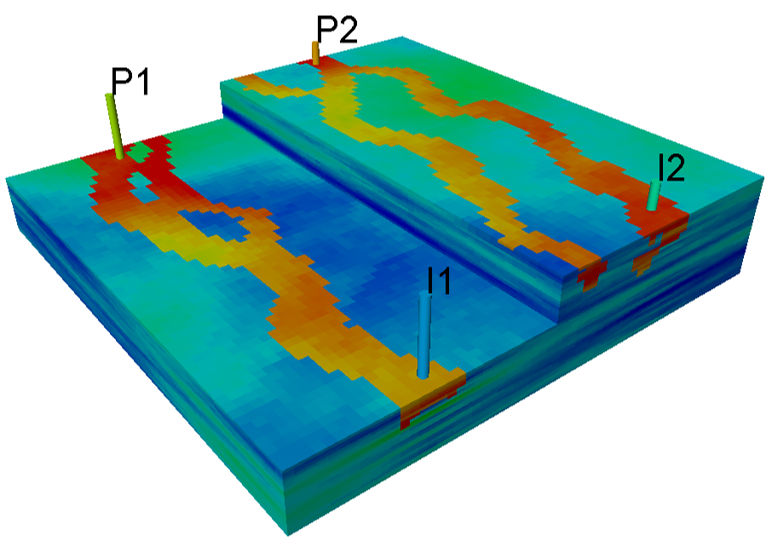}
        \caption{Posterior model \#2}
    \end{subfigure}%
    ~ 
    \begin{subfigure}[b]{0.32\textwidth}
        \includegraphics[width=1\textwidth]{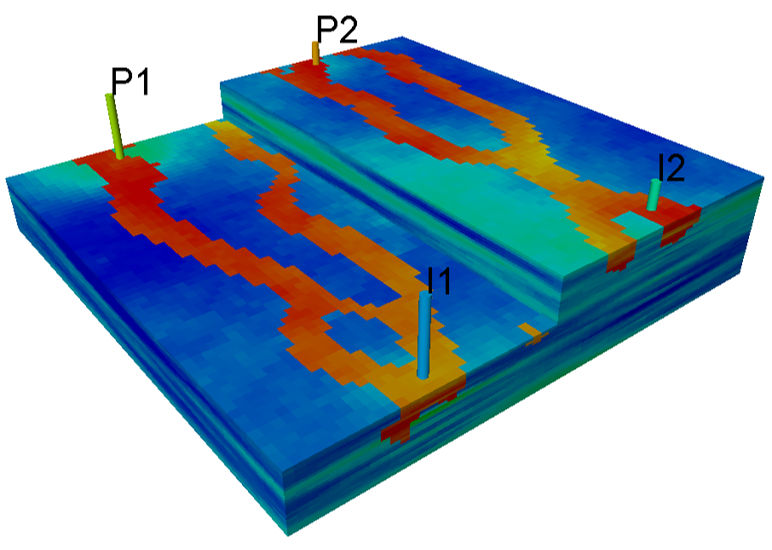}
        \caption{Posterior model \#3}
    \end{subfigure}%
    \vspace{0.2cm}
    \begin{subfigure}[b]{0.6\textwidth}
        \includegraphics[width=1\textwidth]{colorbar.jpg}
    \end{subfigure}%

    \caption{Log-permeability for (a-c) three prior CNN-PCA models and (d-f) corresponding posterior CNN-PCA models. }
    \label{fig_hm_logk}
\end{figure}

Prior and posterior geomodels are shown in Fig.~\ref{fig_hm_logk}. In the true Petrel model (leftmost realization in Fig.~\ref{fig_case3_petrel}), wells I1 and P1, and I2 and P2, are connected via channels. In the first two prior models (Fig.~\ref{fig_hm_logk}a,~b), one or the other of these injector--producer connectivities does not exist, but it gets introduced in the corresponding posterior models (Fig.~\ref{fig_hm_logk}d,~e). The third prior model (Fig.~\ref{fig_hm_logk}c) already displays the correct injector--producer connectivity, and this is indeed retained in the posterior model (Fig.~\ref{fig_hm_logk}f). 


\FloatBarrier

\section{Concluding Remarks}
\label{sec-concl}

In this work, the 3D CNN-PCA algorithm, a deep-learning-based geological parameterization procedure, was developed to treat complex 3D geomodels. The method entails the use of a new supervised-learning-based reconstruction loss and a new style loss based on features of 3D geomodels extracted from the C3D net, a 3D CNN pretrained for video classification. Hard data loss is also included. A two-step treatment for parameterizing bimodal (as opposed to binary) models, involving CNN-PCA representation of facies combined with PCA for within-facies variability, was also introduced.

The 3D CNN-PCA algorithm was applied for the parameterization of realizations from three different geological scenarios. These include a binary fluvial channel system, a bimodal channelized system, and a three-facies channel-levee-mud system. Training required the construction of 3000 object-based Petrel models, the corresponding PCA models, and a set of random PCA models. New PCA models were then fed through the trained model transform net to generate the 3D CNN-PCA representation. The resulting geomodels were shown to exhibit geological features consistent with those in the reference models. Flow results for injection and production quantities, generated by simulating a test set of CNN-PCA models, were found to be in close agreement with simulations using reference Petrel models. Enhanced accuracy relative to truncated-PCA models was also demonstrated. Finally, history matching was performed for the bimodal channel system, using the two-step approach and ESMDA. Significant uncertainty reduction was achieved, and the posterior models were shown to be geologically realistic.

There are a number of directions for future work in this general area. The models considered in this study were Cartesian and contained $60\times60\times40$ cells (144,000 total grid blocks). Practical subsurface flow models commonly contain more cells and may be defined on corner-point or unstructured grids. Extensions to parameterize systems of this type containing, e.g., $O(10^6)$ cells, should be developed. Systems with larger numbers of wells should also be considered. It will additionally be of interest to extend our treatments to handle uncertainty in the geological scenario, with a single network trained to accomplish multi-scenario style transfer. Finally, the development of parameterizations for discrete fracture systems should be addressed.


\section*{Computer Code Availability}
Computer code and datasets will be available upon publication.

\begin{acknowledgements}
We thank the industrial affiliates of the Stanford Smart Fields Consortium for financial support. We are also grateful to the Stanford Center for Computational Earth \& Environmental Science for providing the computing resources used in this work. We also thank Obi Isebor, Wenyue Sun and Meng Tang for useful discussions, and Oleg Volkov for help with the ADGPRS software.
\end{acknowledgements}

\bibliographystyle{spbasic}      
\bibliography{deep_rp,hm_master_thesis,manual,dsi}

\section*{Supplementary Information (SI)}
\setcounter{section}{0}
\beginsupplement
In this SI we first provide details on the architectures of the 3D model transform net and C3D net. Modeling and flow results complementing those in the main text are then presented. These include truncated-PCA results for the three-facies example (Case~2) and PCA results for the bimodal channelized system (Case~3). Visualizations of geomodels and flow statistics are provided for both cases.

\section{3D Model Transform Net Architecture}

The architecture of the 3D model transform net is summarized in Table~\ref{tab-3d-fw}. In the table, `Conv' denotes a 3D convolutional layer immediately followed by a 3D batch normalization layer and a ReLU nonlinear activation. `Residual block' contains a stack of two convolutional layers, each with 128 filters of size $3 \times 3 \times 3 \times 128$ and stride (1, 1, 1). Within each residual block, the first 3D convolutional layer is followed by a 3D batch normalization and a ReLU nonlinear activation. The second convolutional layer is followed only by a 3D batch normalization. The final output of the residual block is the sum of the input to the first 3D convolutional layer and the output from the second 3D convolutional layer. The last three `DeConv' layers consist of a nearest-neighbor upsampling layer, followed by a 3D convolutional layer with stride (1, 1, 1), a 3D batch normalization and a ReLU nonlinear activation. The last `DeConv' layer is an exception as it only contains one 3D convolutional layer. Circular padding is used for all 3D convolutional layers.

\begin{table}[!htb]
\centering
\begin{tabular}{ c | c  }
  \textbf{Layer} & \textbf{Output size} \\
  \hline
  Input & ($\Nz$, $\Ny$, $\Nx$, $1$)  \\
  \\[-1em]
  Conv, 32 filters of size $3 \times 9 \times 9 \times 1$, stride (1, 1, 1) & ($\Nz$, $\Ny$, $\Nx$, $32$)  \\
  \\[-1em]
  Conv, 64 filters of size $3 \times 3 \times 3 \times32$, stride (2, 2, 2) & ($\Nz/2$, $\Ny/2$, $\Nx/2$, $64$)  \\
  \\[-1em]
  Conv, 128 filters of size $3 \times 3 \times 3 \times64$, stride (1, 2, 2) & ($\Nz/2$, $\Ny/4$, $\Nx/4$, $128$)  \\
    \\[-1em]
  Residual block, 128 filters & ($\Nz/2$, $\Ny/4$, $\Nx/4$, $128$)  \\
      \\[-1em]
  Residual block, 128 filters & ($\Nz/2$, $\Ny/4$, $\Nx/4$, $128$)  \\
      \\[-1em]
  Residual block, 128 filters & ($\Nz/2$, $\Ny/4$, $\Nx/4$, $128$)  \\
      \\[-1em]
  Residual block, 128 filters & ($\Nz/2$, $\Ny/4$, $\Nx/4$, $128$)  \\
      \\[-1em]
  Residual block, 128 filters & ($\Nz/2$, $\Ny/4$, $\Nx/4$, $128$)  \\
    \\[-1em]
  DeConv, 64 filters of size $3 \times 3 \times 3 \times 128$, upsample (1, $2$, $2$) & ($\Nz/2$, $\Ny/2$, $\Nx/2$, $64$)  \\
      \\[-1em]
  DeConv, 32 filters of size $3 \times 3 \times 3 \times 64$, upsample ($2$, $2$, $2$) & ($\Nz$, $\Ny$, $\Nx$, $32$)  \\
        \\[-1em]
  Conv, 1 filter of size $3 \times 9 \times 9 \times 64$ & ($\Nz$, $\Ny$, $\Nx$, $1$)  \\
  \hline  
\end{tabular}
\caption{Network architecture used in the 3D model transform net}
\label{tab-3d-fw}
\end{table}

\section{C3D Net Architecture}

The architecture of the C3D transform net is summarized in Table~\ref{tab-3d-c3d}. In the C3D net, each `Conv layer' contains a 3D convolutional layer followed immediately by a ReLU nonlinear activation. The C3D net contains four blocks of convolutional layers. The last convolutional layer at each block, corresponding to layers~1, 2, 4 and 6, is used for extracting features from 3D geomodels. These features are used for evaluating the style loss term. The 3D models are duplicated three times along the feature dimension (the fourth dimension) before being fed to the C3D net.

\begin{table}[!htb]
\centering
\begin{tabular}{ c | c  }
  \textbf{Layer} & \textbf{Output size} \\
  \hline
  Input & ($\Nz$, $\Ny$, $\Nx$, $3$)  \\
  Conv layer 1, 64 filters of size $3 \times 3 \times 3 \times 3$ & ($\Nz$, $\Ny$, $\Nx$, $64$)  \\
  \\[-1em]
  Maxpool layer, kernel size (1, 2, 2), stride (1, 2, 2) & ($\Nz$, $\Ny/2$, $\Nx/2$, $64$)  \\
  \\[-1em]
  \hline
  \\[-1em]
  Conv layer 2, 128 filters of size $3 \times 3 \times 3 \times 64 $ & ($\Nz$, $\Ny/2$, $\Nx/2$, $128$)  \\
  \\[-1em]
  Maxpool layer, kernel size (2, 2, 2), stride (2, 2, 2) & ($\Nz/2$, $\Ny/4$, $\Nx/4$, $128$)  \\
  \\[-1em]
  \hline
  \\[-1em]
  Conv layer 3, 256 filters of size $3 \times 3 \times 3 \times 128 $ & ($\Nz/2$, $\Ny/4$, $\Nx/4$, $256$)  \\
    \\[-1em]
    Conv layer 4, 256 filters of size $3 \times 3 \times 3 \times 256 $ & ($\Nz/2$, $\Ny/4$, $\Nx/4$, $256$)  \\
    \\[-1em]
  Maxpool layer, kernel size (2, 2, 2), stride (2, 2, 2) & ($\Nz/4$, $\Ny/8$, $\Nx/8$, $256$) \\
  \\[-1em]
  \hline
  \\[-1em]
  Conv layer 5, 512 filters of size $3 \times 3 \times 3 \times 256 $ & ($\Nz/4$, $\Ny/8$, $\Nx/8$, $512$)  \\
    \\[-1em]
    Conv layer 6, 512 filters of size $3 \times 3 \times 3 \times 512 $ & ($\Nz/4$, $\Ny/8$, $\Nx/8$, $512$)  \\
    \\[-1em]
  Maxpool layer, kernel size (2, 2, 2), stride (2, 2, 2) & ($\Nz/8$, $\Ny/16$, $\Nx/16$, $512$) \\
  \hline  
\end{tabular}
\caption{Network architecture for the C3D net (excluding the final dense layers)}
\label{tab-3d-c3d}
\end{table}

\section{Truncated-PCA Results for Three-Facies System}

In this section we present truncated-PCA results for the three-facies (channel-levee-mud) system. For truncated PCA, the `natural' encoding (mud denoted by 0, levee by 1, channel by 2) performs significantly better, both in terms of geomodel appearance and flow response, than the alternative encoding used for CNN-PCA. Therefore, here we present truncated-PCA results using this natural encoding. See Section~3.3 in the main paper for a discussion of the alternate CNN-PCA encoding. 

\begin{figure}[!htb]
    \centering
    \begin{subfigure}[b]{0.24\textwidth}
        \includegraphics[width=1\textwidth]{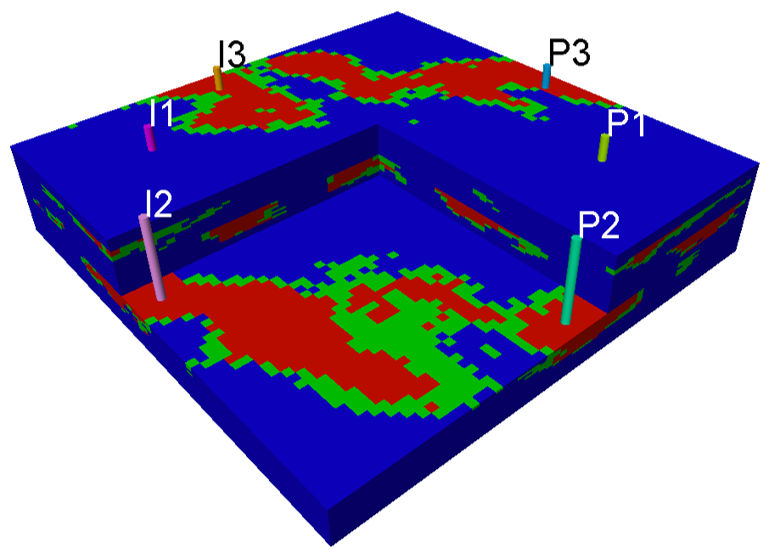}
    \end{subfigure}%
    ~ 
    \begin{subfigure}[b]{0.24\textwidth}
        \includegraphics[width=1\textwidth]{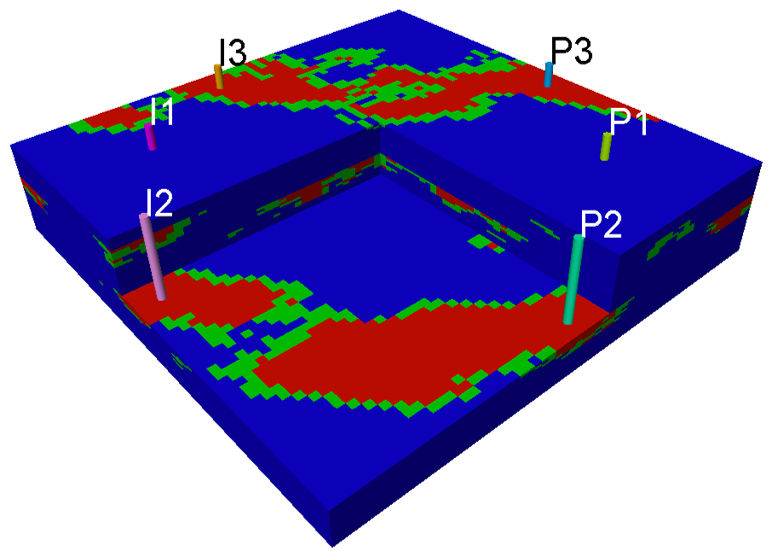}
    \end{subfigure}%
    ~
    \begin{subfigure}[b]{0.24\textwidth}
        \includegraphics[width=1\textwidth]{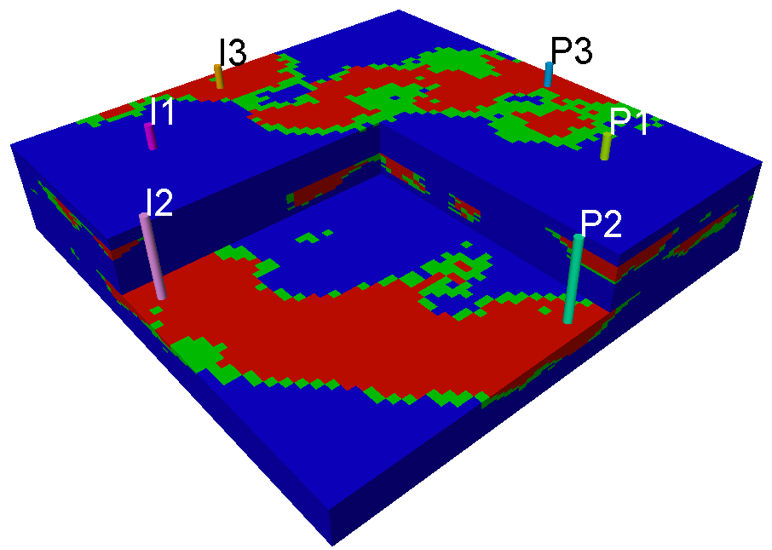}
    \end{subfigure}%
    ~
    \begin{subfigure}[b]{0.24\textwidth}
        \includegraphics[width=1\textwidth]{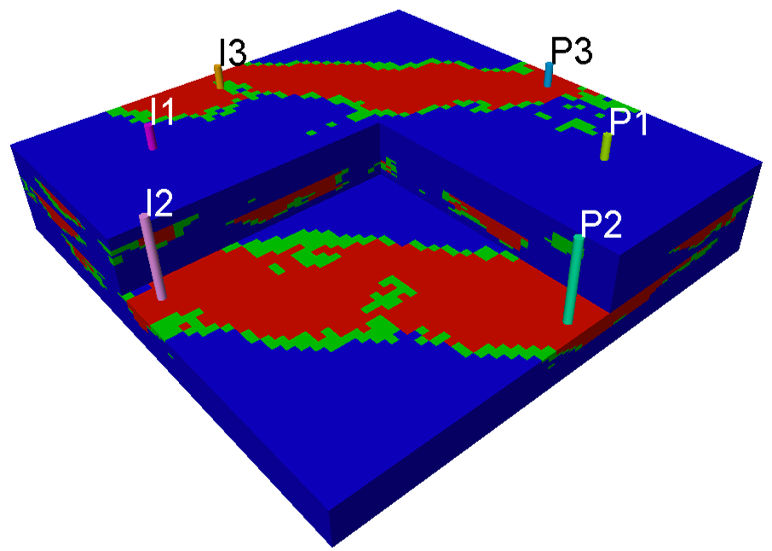}
    \end{subfigure}%

    \caption{Four test-set truncated-PCA realizations of the three-facies system, with sand shown in red, levee in green, and mud in blue (Case~2).}
    \label{fig_case2_tpca_models}
\end{figure}

\begin{figure}[!htb]
    \centering
    \begin{subfigure}[b]{0.32\textwidth}
        \includegraphics[width=1\textwidth]{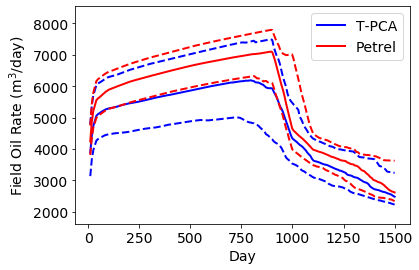}
        \caption{Field oil rate}
    \end{subfigure}%
    ~ 
    \begin{subfigure}[b]{0.32\textwidth}
        \includegraphics[width=1\textwidth]{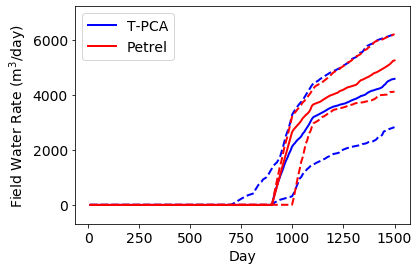}
        \caption{Field water rate}
    \end{subfigure}%
    ~ 
    \begin{subfigure}[b]{0.32\textwidth}
        \includegraphics[width=1\textwidth]{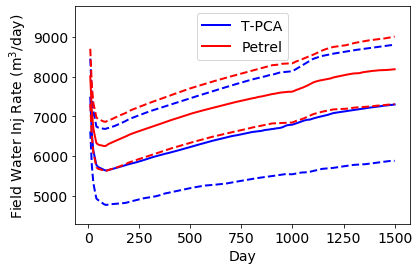}
        \caption{Field water injection rate}
    \end{subfigure}%
    
    \centering
    \begin{subfigure}[b]{0.32\textwidth}
        \includegraphics[width=1\textwidth]{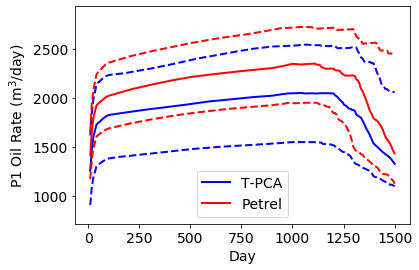}
        \caption{P1 oil rate}
    \end{subfigure}%
    ~ 
    \begin{subfigure}[b]{0.32\textwidth}
        \includegraphics[width=1\textwidth]{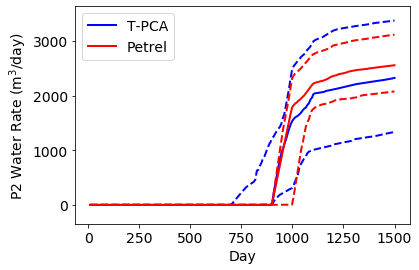}
        \caption{P2 water rate}
    \end{subfigure}%
    ~ 
    \begin{subfigure}[b]{0.32\textwidth}
        \includegraphics[width=1\textwidth]{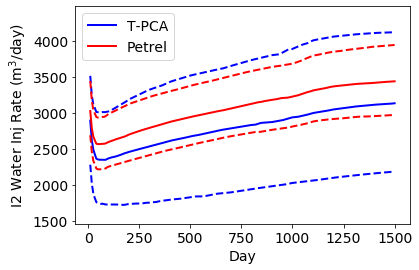}
        \caption{I2 water injection rate}
    \end{subfigure}%
    
    \caption{Comparison of Petrel (red curves) and truncated-PCA (blue curves) flow statistics over ensembles of 200 (new) test cases. Solid curves correspond to $\text{P}_{50}$ results, lower and upper dashed curves to $\text{P}_{10}$ and $\text{P}_{90}$ results (Case~2).}
    \label{fig_case2_flow_stats_tpca}
\end{figure}

The same set of $\Nr=3000$ Petrel realizations (encoded as noted above) is used to construct the PCA representation. The reduced dimension is $l=500$, which captures \textapprox80\% of the total energy. These PCA realizations are truncated such that average facies fractions of mud, levee and channel match those of the Petrel models. Four test-set truncated-PCA models are shown in Fig.~\ref{fig_case2_tpca_models}. With reference to the Petrel realizations (Fig.~10 in the main paper), it is evident that the truncated-PCA geomodels do not display proper channel and levee geometries. Flow results for these models, presented in Fig.~\ref{fig_case2_flow_stats_tpca}, exhibit significant deviation from the reference Petrel flow responses. The CNN-PCA geomodels and flow results (shown in Figs.~11 and 12 in the main paper) are clearly superior to those shown here for truncated PCA.

\section{PCA Results for Bimodal Channelized System}

PCA results for Case~3 in the main paper will now be discussed. Here we construct the PCA representation directly for the bimodal log-permeability field $\Bm$. A total of $\Nr=3000$ Petrel realizations of the log-permeability field are used. For this case we set $l=700$, which explains \textapprox80\% of the total energy. We also generate geomodels for which a histogram transformation is applied to the original PCA realizations, such that the average histogram for the resulting models matches that of the reference Petrel realizations. We refer to the PCA models after histogram transformation as PCA+HT models. 

\begin{figure}[!htb]
    \centering
    \begin{subfigure}[b]{0.24\textwidth}
        \includegraphics[width=1\textwidth]{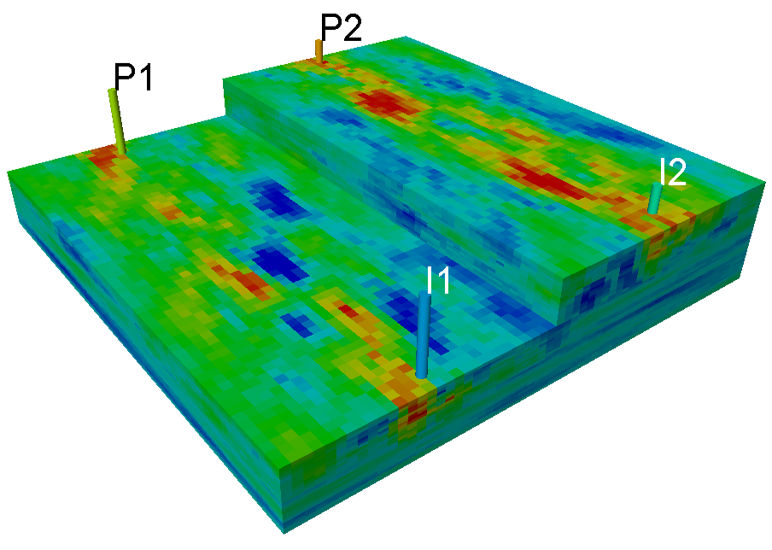}
        \caption{PCA model 1}
    \end{subfigure}%
    ~ 
    \begin{subfigure}[b]{0.24\textwidth}
        \includegraphics[width=1\textwidth]{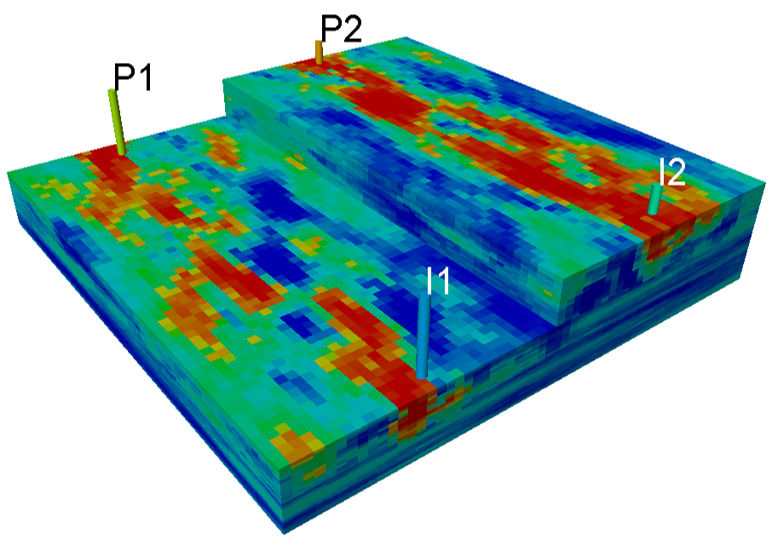}
        \caption{PCA+HT model 1}
    \end{subfigure}%
    ~
    \begin{subfigure}[b]{0.24\textwidth}
        \includegraphics[width=1\textwidth]{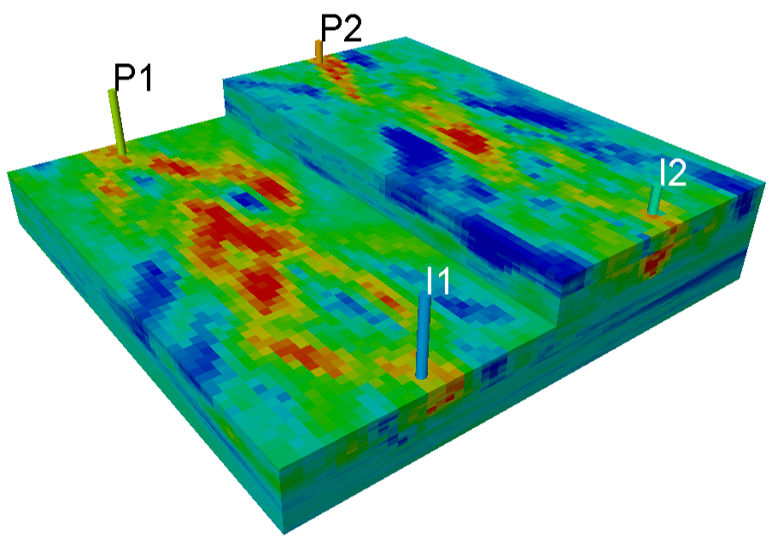}
        \caption{PCA model 2}
    \end{subfigure}%
    ~
    \begin{subfigure}[b]{0.24\textwidth}
        \includegraphics[width=1\textwidth]{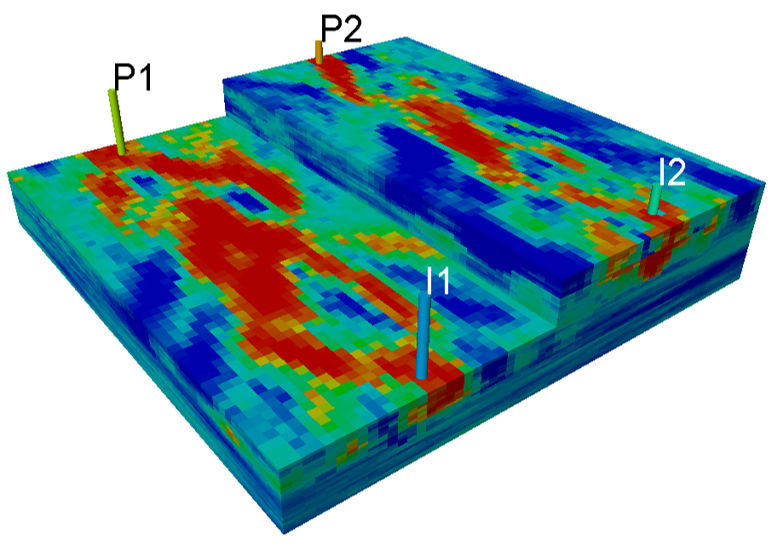}
        \caption{PCA+HT model 2}
    \end{subfigure}%
    \vspace{0.2cm}
    \begin{subfigure}[b]{0.6\textwidth}
        \includegraphics[width=1\textwidth]{colorbar.jpg}
    \end{subfigure}%

    \caption{Two test-set PCA models (a, c) and corresponding PCA+HT models (b, d) for the bimodal channelized system (Case~3).}
    \label{fig_case3_pca}
\end{figure}

Two PCA models, along with the corresponding PCA+HT models, are shown in Fig.~\ref{fig_case3_pca}. By comparing these to the Petrel realizations (shown in Fig.~13 in the main paper), we see that PCA and PCA+HT models do not accurately reproduce the features evident in the reference geomodels. The PCA+HT models do, however, display better channel connectivity than the PCA models. The CNN-PCA geomodels, shown in Fig.~14 in the main paper, more closely resemble the reference realizations.

\begin{figure}[!htb]
    \centering
    \begin{subfigure}[b]{0.32\textwidth}
        \includegraphics[width=1\textwidth]{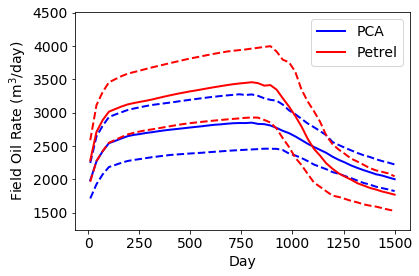}
        \caption{Field oil rate}
    \end{subfigure}%
    ~ 
    \begin{subfigure}[b]{0.32\textwidth}
        \includegraphics[width=1\textwidth]{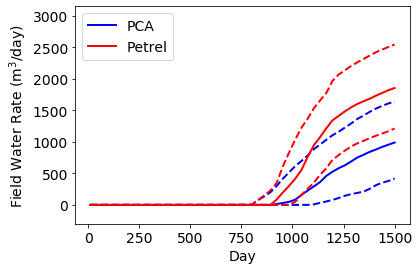}
        \caption{Field water rate}
    \end{subfigure}%
    ~ 
    \begin{subfigure}[b]{0.32\textwidth}
        \includegraphics[width=1\textwidth]{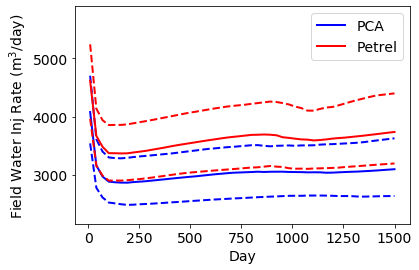}
        \caption{Field water injection rate}
    \end{subfigure}%
    
    \centering
    \begin{subfigure}[b]{0.32\textwidth}
        \includegraphics[width=1\textwidth]{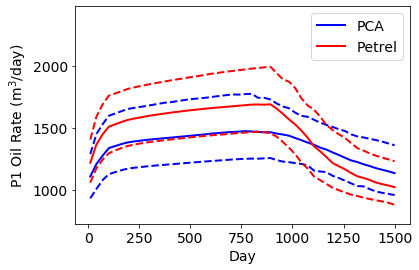}
        \caption{P1 oil rate}
    \end{subfigure}%
    ~ 
    \begin{subfigure}[b]{0.32\textwidth}
        \includegraphics[width=1\textwidth]{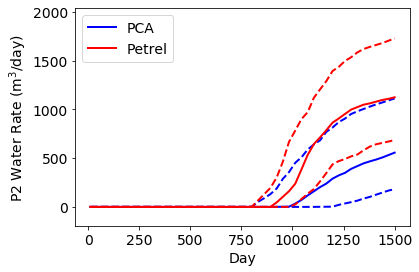}
        \caption{P2 water rate}
    \end{subfigure}%
    ~ 
    \begin{subfigure}[b]{0.32\textwidth}
        \includegraphics[width=1\textwidth]{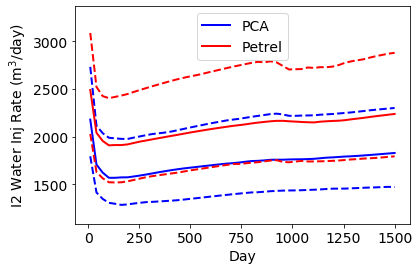}
        \caption{I2 water injection rate}
    \end{subfigure}%

    \caption{Comparison of Petrel (red curves) and PCA (blue curves) flow statistics over ensembles of 200 (new) test cases. Solid curves correspond to $\text{P}_{50}$ results, lower and upper dashed curves to $\text{P}_{10}$ and $\text{P}_{90}$ results (Case~3).}
    \label{fig_case3_flow_stats_pca}
\end{figure}

\begin{figure}[!htb]
    \centering
    \begin{subfigure}[b]{0.32\textwidth}
        \includegraphics[width=1\textwidth]{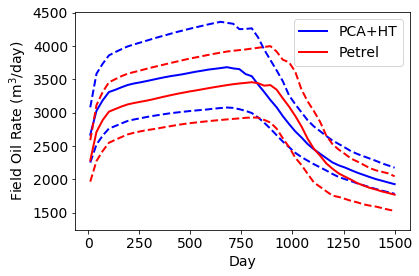}
        \caption{Field oil rate}
    \end{subfigure}%
    ~ 
    \begin{subfigure}[b]{0.32\textwidth}
        \includegraphics[width=1\textwidth]{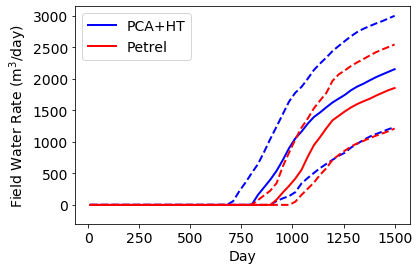}
        \caption{Field water rate}
    \end{subfigure}%
    ~ 
    \begin{subfigure}[b]{0.32\textwidth}
        \includegraphics[width=1\textwidth]{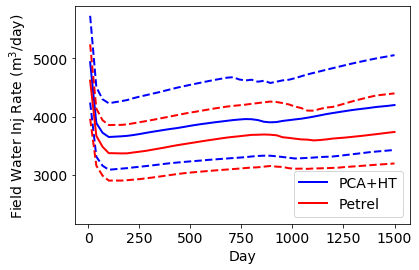}
        \caption{Field water injection rate}
    \end{subfigure}%
    
    \centering
    \begin{subfigure}[b]{0.32\textwidth}
        \includegraphics[width=1\textwidth]{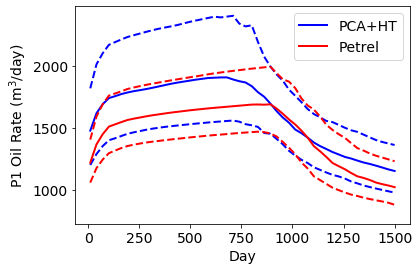}
        \caption{P1 oil rate}
    \end{subfigure}%
    ~ 
    \begin{subfigure}[b]{0.32\textwidth}
        \includegraphics[width=1\textwidth]{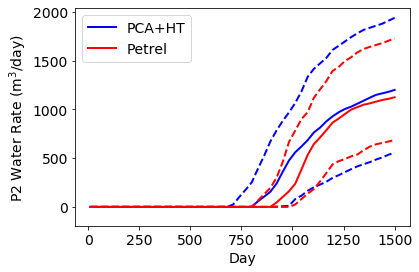}
        \caption{P2 water rate}
    \end{subfigure}%
    ~ 
    \begin{subfigure}[b]{0.32\textwidth}
        \includegraphics[width=1\textwidth]{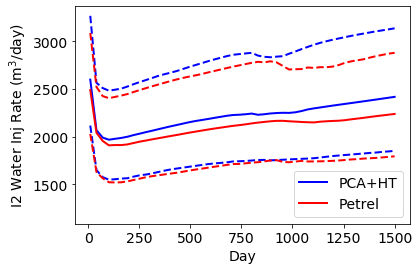}
        \caption{I2 water injection rate}
    \end{subfigure}%

    \caption{Comparison of Petrel (red curves) and PCA+HT (blue curves) flow statistics over ensembles of 200 (new) test cases. Solid curves correspond to $\text{P}_{50}$ results, lower and upper dashed curves to $\text{P}_{10}$ and $\text{P}_{90}$ results (Case~3).}
    \label{fig_case3_flow_stats_pcaht}
\end{figure}

Flow statistics for PCA and PCA+HT geomodels are presented in Figs.~\ref{fig_case3_flow_stats_pca} and \ref{fig_case3_flow_stats_pcaht}, respectively. Interestingly, the PCA models significantly underestimate flow quantities, while the PCA+HT models consistently overestimate these quantities. This may be due to a lack of channel connectivity in the PCA realizations, and overly wide channels in the PCA+HT models. Flow statistics for CNN-PCA geomodels (Fig.~16 in the main paper), by contrast, display much closer overall agreement with the reference results from the Petrel models.

\end{document}